\pdfoutput=1

\documentclass[11pt]{article}

\usepackage[]{EMNLP2023}

\usepackage{times}
\usepackage{latexsym}

\usepackage[T1]{fontenc}

\usepackage[utf8]{inputenc}

\usepackage{microtype}

\usepackage{inconsolata}

\usepackage{hyperref}
\usepackage{xcolor}
\definecolor{darkblue}{rgb}{0, 0, 0.5}
\hypersetup{colorlinks=true,citecolor=darkblue, linkcolor=darkblue, urlcolor=darkblue}

\usepackage{times,latexsym}
\usepackage{url}
\usepackage[T1]{fontenc}

\usepackage{CJKutf8}
\usepackage{xcolor}
\usepackage{adjustbox}
\usepackage{makecell}
\usepackage{multicol}
\usepackage{multirow}
\usepackage{hhline}
\usepackage{subfigure} 
\usepackage{graphicx}
\usepackage{amsmath}
\usepackage{cleveref}
\crefname{section}{§}{§§}
\usepackage{tabularray}
\usepackage{enumitem}
\usepackage{hyperref}
\UseTblrLibrary{counter}
\UseTblrLibrary{functional}
\UseTblrLibrary{booktabs}  
\usepackage{mfirstuc}  
\usepackage{xspace}
\usepackage{tabulary}
\usepackage{soul}
\usepackage{varwidth} 
\usepackage{booktabs}
\sethlcolor{gray!30}

\newcommand{\citeiter}{\citep[][\textit{inter alia}]}
\newcommand{\blank}{\underline{\hspace{0.9em}}}

\definecolor{deeppink}{RGB}{255, 105, 180}
\definecolor{mycolor}{RGB}{240,240,240}
\definecolor{MyGrey}{HTML}{838383}
\definecolor{MyBlue}{HTML}{1F4E79}
\definecolor{MyRed}{HTML}{A80000}
\definecolor{MyYellow}{HTML}{FFCC00}
\definecolor{MyPink}{HTML}{83639f}
\definecolor{MyGreen}{HTML}{449945}
\definecolor{MyOrange}{HTML}{ea7827}
\definecolor{LinkPink}{HTML}{df1a7d}

\definecolor{ForestGreen}{HTML}{009B55}
\definecolor{OrangeRed}{HTML}{c22f2f}
\definecolor{Dandelion}{HTML}{e9963e}

\definecolor{InstanceBlue}{HTML}{6bb2e7}
\definecolor{TaskRed}{HTML}{fe4544}

\newcommand{\MyYes}{\textcolor{ForestGreen}{\textrm{yes}}}
\newcommand{\MyNo}{\textcolor{OrangeRed}{\textrm{no}}}
\newcommand{\MyMaybe}{\textcolor{Dandelion}{\textrm{maybe}}}

\usepackage{dingbat} 
\usepackage{bbding}
 
\newcommand{\entailinstruction}{\textsl{NLI-oriented Instructions}}
\newcommand{\plminstruction}{\textsl{LLM-oriented Instructions}} 
\newcommand{\humaninstruction}{\textsl{Human-oriented Instructions}} 
\newcommand{\TaskDef}{{\color{MyPink}{\em {\normalsize \textrm{Task Definition}}}}}
\newcommand{\Demonstrations}{{\color{MyOrange}{\em {\normalsize \textrm{Demonstrations}}}}}
\newcommand{\TestInput}{{\color{MyGreen}{\em {\normalsize \textrm{Test Instance}}}}}

\usepackage{tikz}
\newcommand\encircle[2][]{\tikz[overlay]\node[fill=blue!20,inner sep=2pt, anchor=text, rectangle, rounded corners=1.5mm,#1] {#2};\phantom{#2}}
\definecolor{YY}{HTML}{E6C802}
\definecolor{myOrangev2}{HTML}{ED8E55}
\definecolor{MyGreenv2}{HTML}{009B55}
\definecolor{MyRedv2}{HTML}{c22f2f}

\newif\iftaclinstructions
\taclinstructionsfalse 
\iftaclinstructions

\newcommand{\instr}
\fi

\title{Large Language Model Instruction Following: A Survey \\ of Progresses and Challenges}
\author{
  Renze Lou\textsuperscript{\rm $\spadesuit$} \quad
  Kai Zhang\textsuperscript{\rm $\diamondsuit$} \quad
  \and
  Wenpeng Yin\textsuperscript{\rm $\spadesuit$}
  \\
  \textsuperscript{\rm $\spadesuit$}The Pennsylvania State University
  \ 
  \textsuperscript{\rm $\diamondsuit$}The Ohio State University
  \\
  {\small \texttt{\{renze.lou, wenpeng\}@psu.edu};}
  \ 
  {\small \texttt{zhang.13253@osu.edu}}
}

\date{}

\begin{document}
\maketitle
\begin{abstract}
Task semantics can be expressed by a set of input-output examples or a piece of textual instruction. Conventional machine learning approaches for natural language processing (NLP) mainly rely on the availability of large-scale sets of task-specific examples. Two issues arise: first, collecting task-specific labeled examples does not apply to scenarios where tasks may be too complicated or costly to annotate, or the system is required to handle a new task immediately; second, this is not user-friendly since end-users are probably more willing to provide task description rather than a set of examples before using the system. 
  Therefore, the community is paying increasing interest in a new supervision-seeking paradigm for NLP: \textit{learning to follow task instructions, i.e., instruction following}. Despite its impressive progress, there are some unsolved research equations that the community struggles with. This survey paper tries to \textit{summarize} and \textit{provide insights} to the current research on instruction following, particularly, by answering the following questions: (i) What is task instruction, and what instruction types exist? (ii) How to model instructions? (iii) What are popular instruction following datasets and evaluation metrics? (iv) What factors influence and explain the instructions' performance? (v) What challenges remain in instruction following? To our knowledge, this is the first comprehensive survey about instruction following.\footnote{The curated paper list can be found at:~{
\hypersetup{urlcolor=LinkPink}\url{https://github.com/RenzeLou/awesome-instruction-learning}}}
\end{abstract}

\begin{figure}[!ht]
 \setlength{\belowcaptionskip}{-12pt}
 \setlength{\abovecaptionskip}{0.6pt}
	\begin{center}
		\centering
		\includegraphics[width=1.02\linewidth]{ 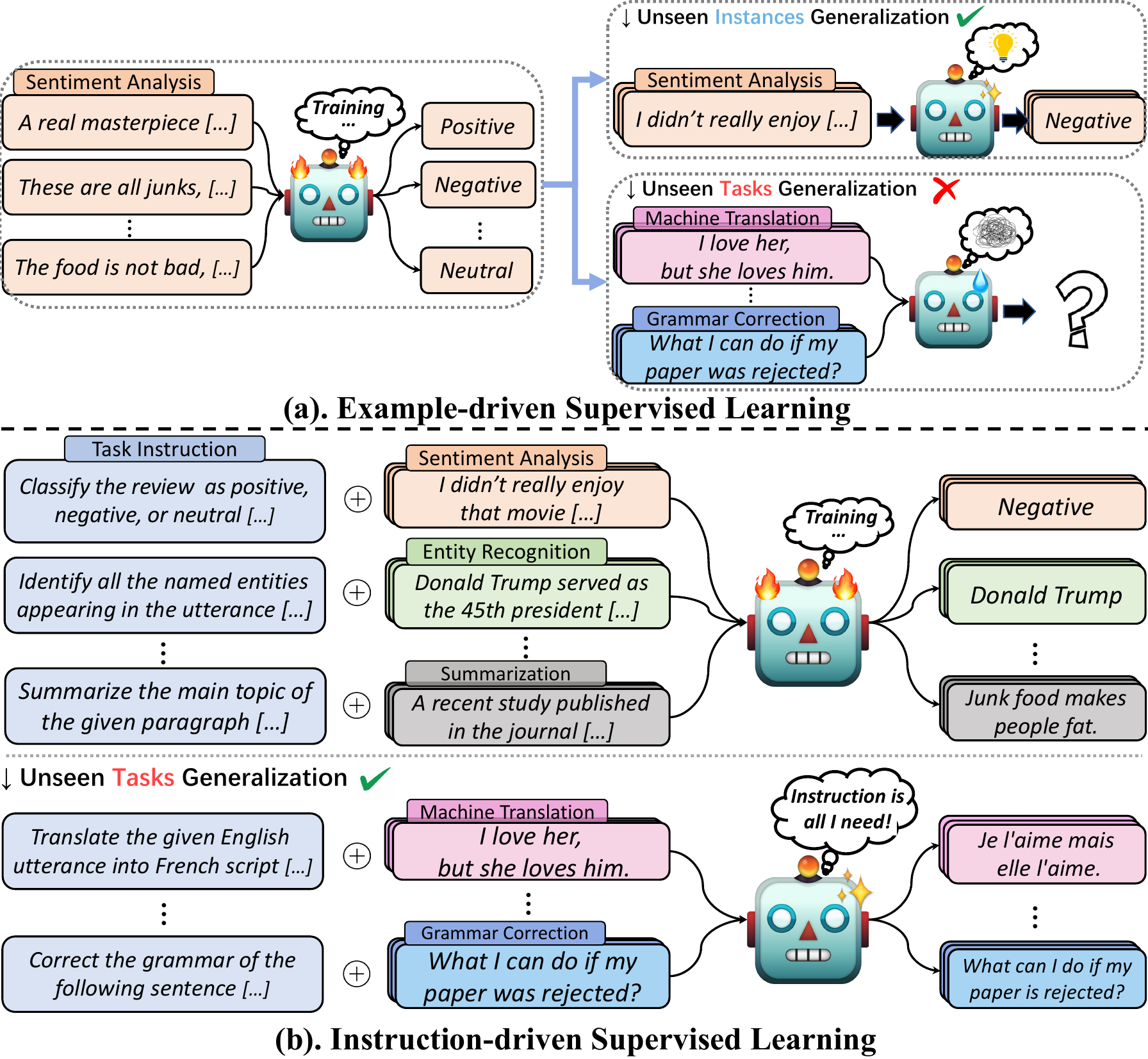}
	\end{center}
	\caption{Two supervised learning paradigms: (a) \textit{example-driven} learning uses extensive labeled examples to represent the task semantics. The resulting system can only generalize to \textcolor{InstanceBlue}{unseen instances} of the same task; (b) \textit{instruction-driven} learning tames model to follow various task instructions. Besides unseen instances, the final system can also generalize to \textcolor{TaskRed}{unseen tasks}.}
	\label{fig:two_paradigms}
\end{figure}

\section{Introduction}

One goal of AI is to build a system that can universally understand and solve new tasks.
Labeled examples (Figure \ref{fig:two_paradigms} (a)), as the mainstream task representation, are costly to obtain at scale or even do not exist in some cases.
Then, is there any other task representation that can contribute to task comprehension?
Textual instructions provide another dimension of supervision for expressing the task semantics, which often contains more abstract and comprehensive knowledge of the target task than individual labeled examples. 
As shown in Figure~\ref{fig:two_paradigms} (b), with the availability of task instructions, systems can be quickly built to handle new tasks.
Such efficiency is highly desirable in real-world applications, especially when task-specific annotations are scarce.
More importantly, instruction following leans toward human intelligence in terms of learning new tasks---a little child can well solve a new mathematical task by learning from its instruction and a few examples~\cite{fennema1996longitudinal, carpenter1996cognitively}.
As a result, this new learning paradigm has recently attracted the main attention of the machine learning and NLP communities~\cite{wang2022benchmarking, longpre2023flan}. 

When talking about ``instruction'', most of us will first think of ``prompt''---using a brief template to convert a task input into a new format (e.g., cloze question) that caters to the language modeling objective of large language models (LLMs)~\cite{brown2020language}.
Despite the prevalence of prompts in text classification, machine translation, etc., we argue that prompts are merely a special case of instructions.
This paper takes a comprehensive and broader view of instruction-driven NLP research. Particularly, we try to answer the following questions: (i) what is task instruction, and what instruction types exist? (\cref{sec:categories}) (ii) given a task instruction, how to encode it to assist the model generalization on the target task? (\cref{sec:modeling}) (iii)  what are popular instruction following datasets and the mainstream evaluation metrics? (\cref{sec:dataset_evaluation}) (iv) what factors (e.g., model size, task numbers) impact the instruction-driven systems' performance? (\cref{sec:analysis}) (v) what challenges exist in instruction following, and what are future directions? (\cref{sec:challenges})

To our knowledge, this is the first paper that surveys the instruction following. In contrast to some existing surveys that focused on a specific in-context instruction, such as prompts~\cite{liu2023pre}, input-by-output demonstrations~\cite{dong2022survey}, or reasoning~\cite{huang2022towards,qiao2022reasoning,yu2023nature}, this work provides a more comprehensive overview of the instruction following.
Our contributions are three-fold:

\textbullet\enspace Going beyond prompts, we analyze prompt constraints via a user-centric lens, with a focus on discerning the disparity between current instruction following research and real-world needs.

\textbullet\enspace We \textit{interpret different task instructions from the unified perspective of \textbf{indirect supervision}}, and summarize their advantages, limitations, and scope of applications;

\textbullet\enspace We regard current ever-growing LLMs and instruction datasets as an effort of dual-track scaling; additionally, we point out current notable research issues and promising directions in the future.

\begin{figure*}[!t]
 \setlength{\belowcaptionskip}{-10pt}
 \setlength{\abovecaptionskip}{5pt}
	\begin{center}
		\centering
        \includegraphics[width=0.95\linewidth]{ 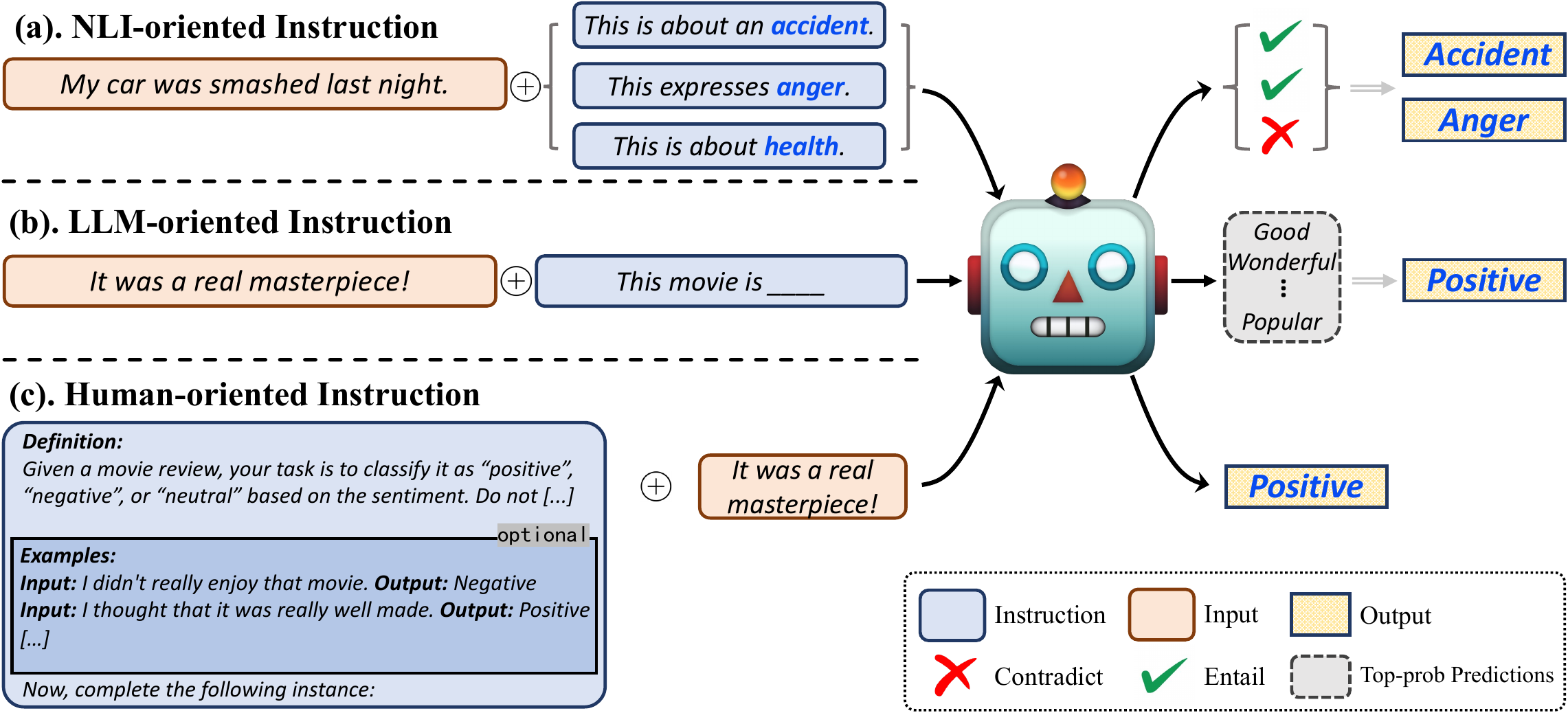}
	\end{center}
	\caption{An illustration of three distinct categories of textual instructions. 
 }
	\label{fig:overview}
\end{figure*}

\section{Related Work}
\label{sec:related}

There are basically two topics that highly relate to this paper, namely \textit{instruction following} (\ref{subsec:instruction_tuning}) and \textit{Surveys on In-context Instructions} (\ref{subsec:related_survey}).

\subsection{Instruction Following}
\label{subsec:instruction_tuning}

As illustrated in Figure~\ref{fig:two_paradigms}, unlike traditional example-driven supervised learning, the essence of instruction following is to train the LLMs to understand various instructions and produce the corresponding responses.
Since this capacity can be extended to any unseen downstream tasks, instruction following has become an efficient learning paradigm for solving few/zero-shot tasks~\citeiter{radford2019language,schick-schutze-2021-just,yin2022contintin,li2023mimic,gupta2023instruction,sun2024umie,xie2024Travel}. However, the performance of instruction following highly relies on both model and task scale: a larger LLM (or pretraining with more tokens) tuned on more diverse tasks can achieve significantly better few/zero-shot performances on the downstream tasks~\citeiter{chung2022scaling,iyer2022opt,wang2023far}. As scaling model size is unrealistic for most of us, numerous recent studies worked on collecting high-quality instruction-tuning datasets, either employing human labors~\cite{khashabi-etal-2020-unifiedqa,ye-etal-2021-crossfit,sanh2021multitask,wang2022benchmarking,longpre2023flan,dolly2023,kopf2023openassistant} or distilling supervision from the powerful LLMs~\cite{wang2022self,honovich2022unnatural,alpaca,peng2023gpt4llm,xu2023baize,koala_blogpost_2023,vicuna2023,xu2023wizardlm,koksal2023longform,kim2023cot,ding2023enhancing,yin2023dynosaur,Lou2023MUFFIN}, e.g., utilizing ChatGPT or GPT-4 to develop creative task instructions~\cite{openai2022chatgpt,OpenAI2023GPT4TR}.

Despite the popularity, the current instruction following still suffers challenges and has considerable room for evolution. This work not only surveys the extensive existing literature on instruction following but also goes beyond: we trace the development of instruction following back to the early days of semantic parsing based machine learning, and formulate our story from an indirect supervision perspective. We hope this survey can systematically introduce this popular yet challenging area.

\subsection{Surveys on In-context Instructions}
\label{subsec:related_survey}

Several existing surveys~\cite{dong2022survey,huang2022towards,qiao2022reasoning,yu2023nature} share similar motivations with us while focusing on merely some sub-area of instruction following, such as prompt, few-shot demonstrations, chain-of-thoughts reasoning, etc.
For example, \citet{liu2023pre} provided a comprehensive overview of prompt learning and LLMs, where the prompt can be regarded as one specific type of textual instruction (as categorized in~\cref{sec:categories}). Some other studies surveying ``soft instruction'', namely parameter-efficient fine-tuning methods~\cite{lialin2023scaling}, also differ from our scope of ``textual instruction''. Notably, \citet{zhang2023instruction} also proposed a survey on instruction tuning, however, they mostly focused on the existing datasets and models; while we present a more complete and consistent story of the instruction following, including the instruction categories, modeling strategies and so on, which previous works have never introduced. To the best of our knowledge, this is the first work that provides a comprehensive and high-level story of instruction following.

\section{Preliminary}
\label{sec:pre}
 
For instruction following, we target driving the systems to reach the corresponding output of the input by following the instruction. Thus, we assume that a dataset usually consists of three items:

\textbullet\enspace \textbf{Input} (\textsc{X}): the input of an instance; it can be a single piece of text (e.g., sentiment classification) or a group of text pieces (e.g., textual entailment, question answering, etc.). 

\textbullet\enspace \textbf{Output} (\textsc{Y}): the output of an instance; in classification problems, it can be one or multiple predefined labels; in text generation tasks, it can be any open-form text. 

\textbullet\enspace \textbf{Template} (\textsc{T}): a textual template that either tries to express task intent or is used for bridging \textsc{X} and \textsc{Y}.\footnote{A plain template connecting \textsc{X} and \textsc{Y}, e.g., ``\texttt{The input is [$\ldots$] The output is [$\ldots$]}'', no task-specific semantics.} \textsc{T} may not be an instruction yet.

In \cref{sec:categories}, we will elaborate that a task instruction \textsc{I} is actually a combination of \textsc{T} with \textsc{X} or \textsc{Y}, or the \textsc{T} on its own in some cases.


\section{What is Task Instruction?---A Unified Perspective from Indirect Supervision}
\label{sec:categories}

This section first summarizes three main instruction types constructed by different combinations of \textsc{T}, \textsc{X}, and \textsc{Y}
(as illustrated in Figure~\ref{fig:overview}), then presents our interpretation of them via an \textit{indirect supervision} perspective.


\begin{table*}[!ht]
    \centering
    \scriptsize
    \resizebox{0.99\textwidth}{!}{
    \begin{tblr}
    {
    width=\linewidth, 
    colspec = {X[l,0.09\linewidth] || X[l,0.36\linewidth] | X[l,0.44\linewidth]},
    rowspec = {Q[b]Q[m]Q[m]Q[m]Q[m]Q[m]},
    row{1} = {bg=azure6, fg=white, font=\bfseries},
    row{even} = {gray!15}, 
    rowhead = 1,
    hspan = minimal,
    }

    Task  & 
    \textsf{NLI~premise (i.e., input text)} &
    \textsf{NLI~hypothesis (i.e., instructions \textsc{Y})} \\ \hline

    \textit{Entity \newline Typing} & [Donald  Trump]$_{ent}$ served as the 45th president of the United States from 2017 to 2021. & (\textcolor{blue}{\checkmark}) Donald  Trump is a \textbf{politician} \newline (\textcolor{red}{\XSolidBrush}) Donald  Trump is a \textbf{journalist}\\\hline

    \textit{Entity \newline Relation}  & [Donald  Trump]$_{ent1}$ served as the 45th president of the [United States]$_{ent2}$ from 2017 to 2021. & (\textcolor{blue}{\checkmark}) Donald  Trump  \textbf{is citizen of} United States \newline (\textcolor{red}{\XSolidBrush}) Donald  Trump  \textbf{is the CEO of} United States \\\hline  

    \textit{Event \newline Argument \newline Extraction} & In [1997]$_{time}$, the [company]$_{sub}$ [hired]$_{trigger}$ [John D. Idol]$_{obj}$ to take over Bill Thang as the new chief executive. & (\textcolor{blue}{\checkmark}) \textbf{John D. Idol}   was hired. \newline (\textcolor{blue}{\checkmark}) \textbf{John D. Idol}   was hired in 1997. \newline (\textcolor{red}{\XSolidBrush}) \textbf{Bill Thang}  was hired. \\\hline

    \textit{\enspace\newline  Event \newline Relation} &  Salesforce  and Slack Technologies  have [entered]$_{event1}$ into a definitive agreement] under which Salesforce will [acquire]$_{event2}$ Slack. & (\textcolor{blue}{\checkmark}) Salesforce acquires Slack \textbf{after} it enters into the agreement with Slack Tech. \newline (\textcolor{red}{\XSolidBrush}) Salesforce acquires Slack \textbf{because} it enters into the agreement with Slack Tech. \\\hline  

    \textit{Stance \newline Detection} & Last Tuesday, Bill said ``animals are equal to human beings'' in his speech. & (\textcolor{blue}{\checkmark}) Bill \textbf{supports} that animals should have lawful rights. \newline (\textcolor{red}{\XSolidBrush}) Bill \textbf{opposes} that animals should have lawful rights.\\\hline

    \end{tblr}
    }
    \caption{{\em{\entailinstruction}}~construct hypotheses to explain the labels (in \textbf{bold}). ``\textcolor{blue}{\checkmark}'':  correct; ``\textcolor{red}{\XSolidBrush}'': incorrect.}
    \label{tab:nlptonli}
\end{table*}

\begin{table*}[!ht]
    \centering
    \scriptsize
    \resizebox{1.0\textwidth}{!}{
    \begin{tblr}
    {
    width=\linewidth, 
    colspec = {X[l,0.08\linewidth] || X[l,0.36\linewidth] | X[l,0.23\linewidth] | X[l,0.07\linewidth] | X[l,0.11\linewidth]},
    rowspec = {Q[b]Q[m]Q[m]Q[m]Q[m]Q[m]},
    row{1} = {bg=azure6, fg=white, font=\bfseries},
    row{even} = {gray!15}, 
    rowhead = 1,
    hspan = minimal,
    }
    \SetCell[c=1]{c} Task & \SetCell[c=1]{c} Input~\textsc{X} & \SetCell[c=1]{c} Template~\textsc{T} (cloze question) & \SetCell[c=1]{c} Answer & \SetCell[c=1]{c} Output~\textsc{Y} \\\hline

    
    \textit{Sentiment Analysis} & I would like to buy it again. & \texttt{[\textsc{X}]} The product is \blank. & Great \newline Wonderful \newline $\ldots$ & Positive \\\hline

    \textit{Entity Tagging} & [Donald Trump]$_{ent}$ served as the 45th president of the United States from 2017 to 2021. & The entity in \texttt{[\textsc{X}]} is a \blank class? & Politician \newline President \newline $\ldots$ & People \\\hline

    \textit{Relation Tagging} & [Donald  Trump]$_{ent1}$ served as the 45th president of the [United States]$_{ent2}$ from 2017 to 2021. & \texttt{[\textsc{X}]} entity$_1$ is the \blank of entity$_2$? &  Executive \newline Leader \newline $\ldots$ & President \\\hline

    \textit{Textual Entailment} & \texttt{[\textsc{X$_1$}]}: Donald Trump served as the 45th president of the United States from 2017 to 2021. \newline \texttt{[\textsc{X$_2$}]}: Donald Trump is a citizen of United States. & \texttt{[\textsc{X$_2$}]}? \blank, because \texttt{[\textsc{X$_1$}]} & Indeed \newline Sure \newline $\ldots$ & Yes \\\hline

    \textit{Translation} & Donald Trump served as the 45th president of the United States from 2017 to 2021. & Translate \texttt{[\textsc{X}]} to French:~\blank &  / & été président~{...} \\\hline
    
    \end{tblr}
    }
    \caption{{\em \plminstruction}~utilize templates to convert the origin inputs into fill-in-blank questions. In most classification tasks, the intermediate answers may require further mapping (i.e., verbalizer).}
    \label{tab:LLM-orentied}
\end{table*}

\subsection{Three Types of Instructions}

\subsubsection{\entailinstruction~(i.e., \textsc{I}=\textsc{T}+\textsc{Y})}

A conventional scheme to handle the classification tasks is to convert the target labels into indices and let models decide which indices the inputs belong to. This paradigm only encodes the input semantics while losing the label semantics. To let systems recognize new labels without relying on massive labeled examples, \citet{yin2019benchmarking} proposed converting the target classification tasks into natural language inference (NLI) by building a hypothesis for each label---deriving the truth value of a label is then converted into determining the truth value of the hypothesis.  As exemplified in Figure~\ref{fig:overview}~(a), this approach builds instructions (\textsc{I}) by combining a template (\textsc{T}) with a label (\textsc{Y}) to explain the task semantics. Table~\ref{tab:nlptonli} further provides more detailed examples for~\entailinstruction.

The advantages of {\entailinstruction} learning are four-fold: (i) it keeps the label semantics and makes it possible to encode the input-output relations; (ii) it unifies various classification problems into an NLI task; (iii) by making use of the indirect supervision from existing NLI datasets, a model trained on NLI tasks is expected to work on other tasks in a zero-shot manner; (iv) it extends the original close-set indices classification problem into an open-domain label recognition paradigm. Therefore, it has been widely used in a variety of few/zero-shot classification tasks~\cite{xu2022universal}, such as classifying topics~\cite{yin2019benchmarking}, sentiments~\cite{zhong2021adapting}, stances~\cite{xu-etal-2022-openstance}, entity types~\cite{li2022ultra}, entity relations~\cite{murty2020expbert,xia2021incremental,sainz-etal-2021-label,sainz-etal-2022-textual}, etc.

\begin{table*}[!ht]
    \centering
    \footnotesize
    \resizebox{1.0\textwidth}{!}{
    \begin{tblr}
    {
    width=\linewidth, 
    colspec = {X[l,0.08\linewidth] || X[l,0.23\linewidth] | X[l,0.73\linewidth] | X[l,0.08\linewidth]},
    rowspec = {Q[b]Q[m]Q[m]Q[m]},
    row{1} = {bg=azure6, fg=white, font=\bfseries},
    row{even} = {gray!15}, 
    rowhead = 1,
    hspan = minimal,
    }

    \SetCell[c=1]{c} Task  & \SetCell[c=1]{c} Input~\textsc{X} & \SetCell[c=1]{c} Template~\textsc{T} + Few-shot Demonstrations & \SetCell[c=1]{c} Output~\textsc{Y}  \\\hline

    \textit{Sentiment \newline Analysis} & I am extremely impressed with its good performance. I would like to buy it again! 
     & 
    \TaskDef: \newline
    \colorbox{MyPink!20}{\quad In this task, you are given a product review, and you need to identify~$\dots$} \newline
    \Demonstrations~\textsf{(optional)}: \newline
    \colorbox{MyOrange!15}{\quad Input: \textsl{These are junks, I am really regret...} \quad Output: \textsl{Negative}} \newline
    \colorbox{MyOrange!15}{\quad Input: \textsl{Wonderful bulb with good duration...} \quad Output: \textsl{Positive}} \newline
    \TestInput: \newline
    \colorbox{MyGreen!20}{\quad Input: \texttt{[\textsc{X}]} \quad Output:~\blank}
     & 
     Positive\\\hline

    \textit{Named \newline Entity \newline Extraction} & Donald Trump served as the 45th president of the United States from 2017 to 2021. 
    &
    \TaskDef: \newline
    \colorbox{MyPink!15}{\quad Your task is to recognize the name of a person in the given sentence~$\dots$} \newline
    \Demonstrations~\textsf{(optional)}: \newline
    \colorbox{MyOrange!15}{\quad Input: \textsl{Ousted WeWork founder Adam Neuman...} \quad Output: \textsl{Adam Neuman}} \newline
    \colorbox{MyOrange!15}{\quad Input: \textsl{Tim Cook became the CEO of Apple Inc since...} \quad Output: \textsl{Tim Cook}} \newline
    \TestInput: \newline
    \colorbox{MyGreen!20}{\quad Input: \texttt{[\textsc{X}]} \quad Output:~\blank}
    & 
    Donald Trump \\\hline
    
    \end{tblr}
    }
    \caption{Two examples that illustrate the {\em \humaninstruction}~(w/ 2-shot demonstrations). Similar to the \plminstruction, \humaninstruction~use task-level templates to convert the origin inputs into blank questions. However, the templates here have sufficient task semantics (i.e., {\color{MyPink}{\em \textrm{Task Definition}}}) and are sometimes equipped with~{\color{MyOrange}{\em \textrm{Demonstrations}}}, while those in \plminstruction~usually do not.}
    \label{tab:human-orentied}
\end{table*}

\subsubsection{\plminstruction~(i.e., prompts; \textsc{I}=\textsc{T}+\textsc{X})}

As shown in Figure~\ref{fig:overview}~(b) and Table~\ref{tab:LLM-orentied}, the prompt is a representative of the~\plminstruction, which is usually a brief utterance prepended with the task input (prefix prompt), or a cloze-question template (cloze prompt). It is basically designed for querying the intermedia responses (that can be further converted into the final outputs) from the LLM.
Since the prompted input conforms to the pre-training objectives of LLM (e.g., the cloze-style input satisfies the masked language modeling objective~\cite{kenton2019bert}), it helps get rid of the reliance on the traditional supervised fine-tuning and greatly alleviates the cost of human annotations. Thus, prompt learning achieved impressive results on a multitude of previous few/zero-shot NLP tasks, like question answering~\cite{radford2019language,lin2021few}, machine translation~\cite{li-etal-2022-prompt}, sentiment analysis~\cite{wu-shi-2022-adversarial}, textual entailment~\cite{schick2021exploiting,schick2021few}, entity recognition~\cite{cui2021template,wang2022instructionner}, etc.

Despite the excellent performance of prompt techniques, there are still two obvious shortcomings with \plminstruction~in real-world applications. (i) \textit{Not User-Friendly}. As the prompt is crafted for serving LLMs, it is encouraged to design the prompt in a ``model's language'' (e.g., model-preferred incoherent words or internal embedding). However, this LLM-oriented style is hard to be understood by users and often violates human intuitions~\cite{gao-etal-2021-making,li-liang-2021-prefix,qin-eisner-2021-learning,khashabi2022prompt}. Meanwhile, the performance of prompts highly depends on the laborious prompt engineering~\cite{bach2022promptsource}, but most end-users are not LLM experts and usually lack sufficient knowledge to tune an effective prompt. (ii) \textit{Applications Constraints}. The prompt is usually short and simplistic, whereas many tasks cannot be effectively formulated with solely a brief prompt, making prompt hard to deal with the diverse formats of real-world NLP tasks~\cite{chen2022knowprompt,zhang2023aligning}.

\subsubsection{\humaninstruction~(i.e., \textsc{I}=\textsc{T}+ optional $\{\textsc{X}_i,\textsc{Y}_i\}_{i=1}^k$)}

\humaninstruction~essentially denotes the instructions used for crowd-sourcing on the human-annotation platforms (e.g., Amazon MTurk). Unlike \plminstruction,  \humaninstruction~ (Figure~\ref{fig:overview}~(c)) are usually some human-readable, descriptive, and paragraph-style information consisting of various components, such as ``\texttt{task title}'', ``\texttt{category}'', ``\texttt{definition}'', and ``\texttt{things to avoid}'', etc.~\citep[cf.][]{mishra2022cross} Thus, \humaninstruction~are more user-friendly and can be ideally applied to almost any complex NLP task. Table~\ref{tab:human-orentied} further shows some representative task examples.

%
Accordingly, \humaninstruction~have attracted much more attention in recent years~\citeiter{hu2022context,gupta-etal-2022-instructdial,yin2022contintin}. However, due to the complex nature, \humaninstruction~are more challenging to encode by vanilla LLMs. For example, off-the-shelf GPT-2 was found to work poorly on following MTurk instructions~\cite{wolf2019huggingface,efrat2020turking}. 
To tame the LLMs better understand the \humaninstruction, follow-up works began to collect large-scale instruction datasets~\cite{mishra2022cross,wang2022benchmarking}. All the previous results showed that, after fine-tuning with various task instructions, the text-to-text LLMs, like T5~\cite{raffel2020exploring}, OPT~\cite{zhang2022opt} and Llama~\cite{touvron2023llama}, achieved remarkable few/zero-shot generalizations by following these complex instructions~\cite{wang2023far,ivison2023camels}.

\begin{table*}[t]
 \setlength{\belowcaptionskip}{-10pt}
 \setlength{\abovecaptionskip}{5pt}
    \centering
    \scriptsize
    \resizebox{1.01\textwidth}{!}{
    \begin{tblr}
    {
        width=\linewidth, 
        colspec = 
            {
            X[l,0.34\linewidth]
            X[c,0.18\linewidth]
            X[c,0.17\linewidth]
            X[c,0.20\linewidth]
            },
        rowspec = 
            {
            |[2pt,MyGrey]Q[b]
            |[1pt,MyGrey,solid]Q[m]Q[m]Q[m]Q[m]
            |[0.8pt,MyGrey,dashed]Q[m]Q[m]Q[m]
            Q[m]Q[m]|[2pt,MyGrey]
            },
        row{1} = {font=\bfseries},
        rowhead = 1,
        hspan = minimal,
    }
    {\footnotesize Trait} & {\footnotesize \texttt{NLI-oriented}} & {\footnotesize \texttt{LLM-oriented}} & {\footnotesize \texttt{Human-oriented}} \\
    Update LLM parameter? & \MyYes & \MyMaybe & \MyYes \\ 
    Require  super large LLMs? & \MyNo & \MyYes & \MyNo \\
    Require further label mapping (e.g., verbalizer)? & \MyYes & \MyYes & \MyNo \\
    End-user friendly? & \MyNo & \MyNo & \MyYes \\
    Instruction granularity & sentence-level (brief) & sentence-level (brief) & paragraph-level (complex) \\
    Instruction scope & output-wise & input-wise & task-wise \\
    Task scope & classification & classification \& generation & classification \& generation \\
    Modeling objective  & NLI  & language modeling & follow instructions \\
    Source of indirect supervision & NLI  & language modeling  & various Text-to-Text tasks \\

    \end{tblr}
    }
    \caption{Comparison of the three different instruction types in~\cref{sec:categories}.}
    \label{tab:comparison}
\end{table*}

\subsection{An Indirect Supervision Perspective}
Although the three types of instructions are very different from each other, they are essentially seeking the same thing---\textit{indirect supervision}---to cope with target tasks that have limited annotations. 

Specifically, \entailinstruction~transform target NLP problems into a source task---NLI---so that the rich supervision from existing NLI datasets can act as indirect supervision for those target problems. \plminstruction~reformat target problems into the source task---language modeling, so that the rich generic-purpose knowledge in those LLMs can be directly utilized to get the output. Whether it is \entailinstruction~or \plminstruction, both try to solve unseen tasks with a generalizable system. However, both of them have limited application scope, e.g., they cannot efficiently deal with some structured prediction tasks~\cite{chen2022knowprompt,zhang2023aligning}. Instead of seeking supervision from a single source task (NLI or language modeling), \humaninstruction~learn indirect supervision from a large set of training tasks, the resulting system, therefore, can ideally generalize to any unseen textual tasks.
\Cref{tab:comparison} further compares them from different dimensions.

\section{How to Model Instructions?}
\label{sec:modeling}

Since both \entailinstruction~and \plminstruction~are associated with either the input $\textsc{X}$ or the output $\textsc{Y}$, these types of instructions do not require specific system design to encode them. \entailinstruction~can be handled by regular systems for the NLI task, and \plminstruction~are mostly fed to auto-regressive LLMs. In contrast, \humaninstruction~are the most challenging type since it is independent of any labeled instances. 

Therefore, this section mainly presents several mainstream modeling strategies for the \humaninstruction, as illustrated in Figure~\ref{fig:modeling_strategies}.


\subsection{Semantic Parser}

\begin{figure}[!t]
 \setlength{\belowcaptionskip}{-9pt}
 \setlength{\abovecaptionskip}{5pt}
	\begin{center}
		\centering
		\includegraphics[width=1.03\linewidth]{ 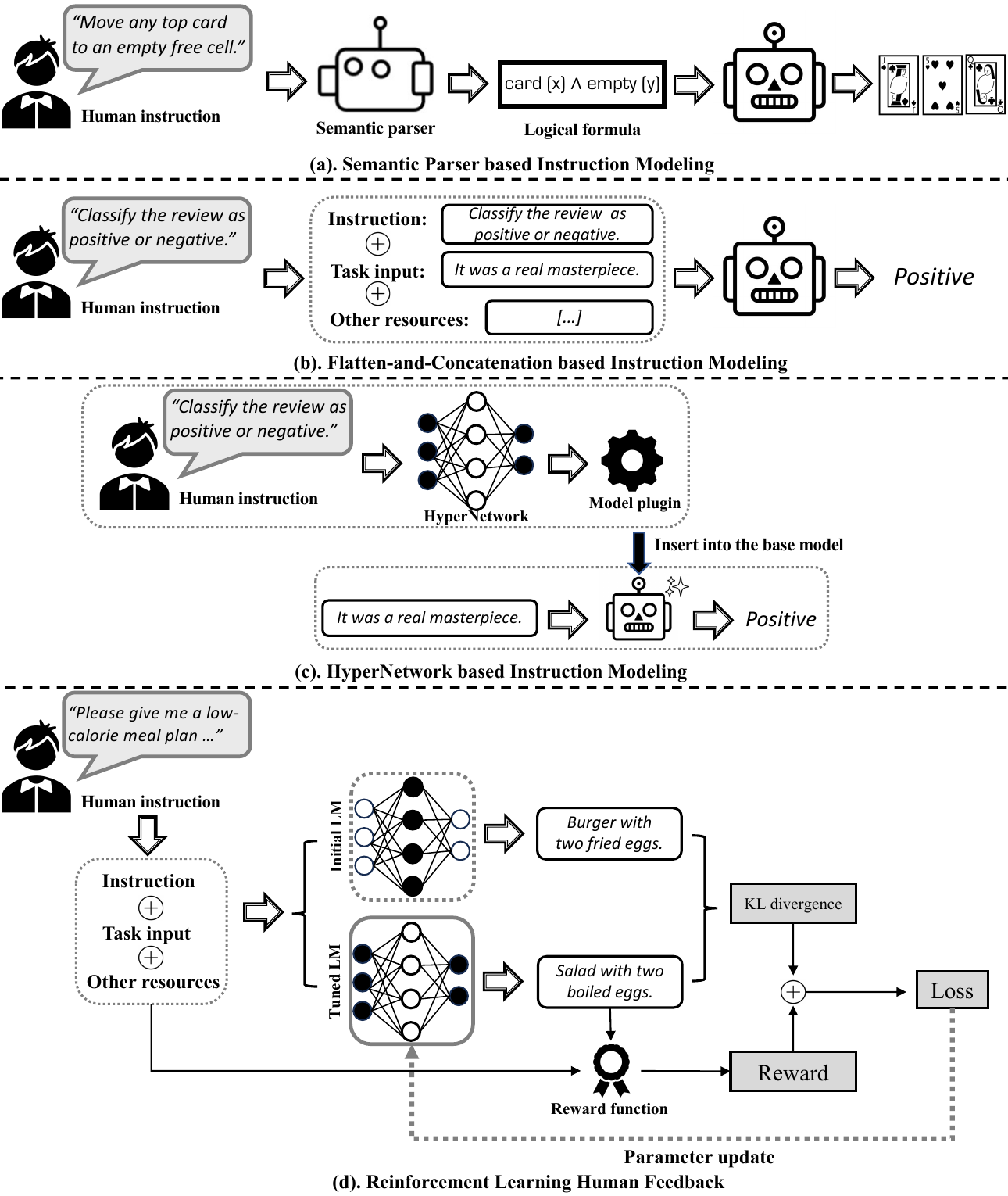}
	\end{center}
	\caption{Four modeling strategies for instructions.
 }
	\label{fig:modeling_strategies}
\end{figure}

At the early stage of machine learning, to help the systems understand natural language instructions, a great number of works employed semantic parsing to convert the instruction into the formal language (logical formula), which can be more easily executed by the systems~\citeiter{goldwasser2014learning}. As exemplified in Figure~\ref{fig:modeling_strategies}~(a), a game instruction ``\texttt{Move any top card to an empty free cell}'' can be processed into an executable formula: ``\texttt{card(x) $\wedge$ freecell(y)}''.

Previous research spent extensive efforts on this strategy, among which most are used for human-computer interaction tasks, e.g., playing soccer games~\cite{kuhlmann2004guiding}. To alleviate laborious human annotations, the follow-up works leveraged indirect or weak supervision from the grounded environments (e.g., knowledge base) to train the semantic parser~\cite{kim2012unsupervised}.



\paragraph{Limitations} Semantic parser-based approaches mainly apply to individual tasks rather than universal cross-task generalization, because building a versatile semantic parser for all NLP tasks is over challenging. By contrast, the approach introduced in the next subsection aims at cross-task generalization with limited supervision for the target tasks.


\subsection{Flatten-and-Concatenation}
\label{subsec:flatten-concatenation}

In contrast to the semantic parser approach, which considers the instructions' structure and the target problems, methods based on the neural networks take more brutal treatment: as illustrated in Figure~\ref{fig:modeling_strategies}~(b)---instructions, regardless of their length, structure, task types, etc., are flattened as a long token sequence and concatenated with the input $\textsc{X}$ as a new input sequence for the models, which has been widely adopted by the prior research~\citeiter{wang2022benchmarking,wei2023symbol}. However, this naive strategy constantly results in unsatisfactory performances when using vanilla models~\cite{weller2020learning}, leading to its reliance on large-scale instruction fine-tuning, well-known as ``instruction tuning''.

\paragraph{Limitations} (i) Flattening and concatenating everything into a long sequence tends to ignore some key information that humans can often capture in the instruction~\cite{mishra-etal-2022-reframing,jang2022can}, such as negation (e.g., ``{{\fontfamily{lmtt}\selectfont do not generate outputs longer than 5 tokens}}''), warning (e.g., ``{\fontfamily{lmtt}\selectfont generate `D' if the question is not answerable or you're not sure}''), output constraints (e.g., ``{\fontfamily{lmtt}\selectfont your answer should be in one of `A', `B', `C', and `D'}''), and so on. (ii) To let models understand the instruction, a large number of training tasks have to be prepared. This is similar to what happened in the early years of deep learning in NLP: to improve the performance of deep neural networks for a particular task, we collect more labeled examples; back to the instruction following, the system's comprehension of the instruction, unfortunately, still exhibits a high degree of dependence on  the scale of training tasks~\cite{chung2022scaling}.

\subsection{HyperNetwork}



Unlike the conventional modeling strategy that encodes the input sequence into the dense representation (i.e., language-to-representation), hypernetwork follows a language-to-parameter paradigm: as shown in Figure~\ref{fig:modeling_strategies}~(c), this scheme converts textual instruction into a block of model parameters that can be further plugged into the underlying models~\cite{ha2016hypernetworks,houlsby2019parameter,jin2020language}. As a result, hypernetwork-based instruction modeling can better leverage the structured input sequence by encoding the instruction and task input separately (i.e., instruction-to-parameter, input-to-representation), achieving stronger generalization compared with the flatten-and-concatenation approach~\cite{ye2021learning,deb2022boosting}. It can also significantly improve inference efficiency, as concluded by recent works~\cite{ivison2022hint}.

\paragraph{Limitations} 
Despite the attractive attributes of hypernetwork, its training instability and the reliance on architecture design (suiting the underlying models) are the stumbling blocks in real-world applications~\cite{Brock2017SMASHOM,ortiz2023non}.

\subsection{Reinforcement Learning from Human Feedback}

The loss function for training LMs significantly impacts the resulting LMs' instruction-following performance~\cite{tay2022unifying}. However, almost all the aforementioned modeling strategies (except the semantic-parser-based method) adopt a conventional next token prediction loss (e.g. cross entropy) to train the models, which tries to capture the human preference by simply comparing the model's generation text with the ground truth reference. 
In order to directly optimize the LMs with the supervision of human preference, recent works utilized reinforcement learning from human feedback (RLHF) to train the LMs~\cite{stiennon2020learning,bai2022training,ouyang2022training}. 

\paragraph{Initial and tuned LM}

The first step of RLHF is to obtain an initial LM, which is usually trained with the flatten-and-concatenation-based modeling strategy --- concatenate instruction, input and all other resources (if they exist) into one input sequence, and train the LM to generate the ground-truth output (as we have introduced before). With the initial LM as a starting point, we can copy it to another independent parameter, which is the target LM will be continually updated in RLHF (i.e., tuned LM).

\paragraph{Prediction shift penalty}

As shown in Figure~\ref{fig:modeling_strategies}~(d), given the initial LM and its copy (the target tuned LM), for each input sequence (e.g., instruction and task input), these two different LMs will generate two outputs. After getting the generation texts of the initial and tuned LM, we can further calculate the textual difference penalty between them:

\begin{align*}
    y &= \theta(I, x) \\
    y^* &= \theta^*(I, x) \\
    r_\textrm{KL} &= \textrm{KL}(y, y^*)
\end{align*}

Here, the \(\theta\) and \(\theta^*\) represent the parameters of initial and tuned LM, respectively. \(I\) and \(x\) denote the instruction and task input; While \(y\) and \(y^*\) are the outputs of the initial and tuned LM. \(r_\textrm{KL}\) is the final reward (loss) of prediction shifting, and the \(\textrm{KL}\) means the calculation of Kullback–Leibler (KL) divergence.
KL divergence is a widely adopted strategy for measuring the textual difference, which can be used as a part of the loss to penalize the tuned LM on shifting the output substantially away from the initial LM's generation. This prediction shift penalty can prevent the tuned LM from fooling the reward function to get a high reward but loses the coherence of the generation text.

\paragraph{Reward function}

Besides the prediction shift penalty, another part of the final reward comes from the reward function. A reward function is used for directly measuring how well the model's output aligns with human preference --- the higher rewards mean the output better aligns with human preference.

A reward function is another model tuned on human preference data, which is usually smaller than the initial LM. It receives the instruction, task input and models' outputs, and tries to predict a reward scalar to reflect the alignment:

\[
r_\textrm{alignment} = r_\theta(I, x, y^*)
\]

The \(r_\theta\) is the reward model, which is usually a regression model.

For training a reward model, previous works collected a bunch of human preference data. For example, \citet{bai2022training} collected different outputs for each input instruction, and asked human annotators to decide which output more aligned with their preference. The training loss of reward model can be formulated as follows:

\[
l=-\log \left(\sigma\left(r_\theta\left(I, x, y^{+}\right)-r_\theta\left(I, x, y^{-}\right)\right)\right)
\]

Where the \(\sigma\) is the activation function that scales the reward into $(0, 1]$, \(y^{+}\) is the output preferred by human, \(y^{-}\) otherwise. By training on these pairwise preference comparison data, the reward model can directly learn to capture the human preference and make alignment reward estimation for the RLHF.

\paragraph{The final training reward and inference}

To this end, the final reward for RL update rule is:

\[
r=r_\textrm{alignment}-\lambda r_{\textrm{KL}}
\]

The \(\lambda\) here is the controlling factor. 

After training with the above RL policy, the final tuned LM can better align with human preference. While the inference procedure of the tuned LM is actually similar to the aforementioned flatten-and-concatenation modeling, where it receives instruction and input, and then generates the corresponding output.

\paragraph{Limitations} 
Compared with other modeling strategies, RLHF requires much more expensive human efforts because of collecting the preference data, especially when the preference comparison outputs are all written by humans~\cite{ouyang2022training}. Meanwhile, the performance of RLHF highly relies on the quality of its human preference annotations. More importantly, in some cases, such as some open-ended creative writing tasks, different humans often hold high disagreement on the preference decision due to the lack of ground-truth output.

\section{Instruction Following Dataset and Evaluation}
\label{sec:dataset_evaluation}

In this section, we shed light on an important topic related to the instruction following, i.e., the instruction-following datasets and the evaluation settings for the instruction-tuned models.

\subsection{Dataset}

The essence of instruction following is to tame the models by following various task instructions and responding with the corresponding desired outputs. Therefore, the instruction-tuning dataset (high-quality instruction-output pairs) is the critical part~\cite{wang2023far,zhou2023lima}.
The current instruction-tuning datasets can be divided into two categories according to different annotation categories: 1) \textbf{human-annotated datasets}~(\cref{subsubsec:human_datasets}); 2) \textbf{LLM-synthetic datasets}~(\cref{subsubsec:LLM_datasets}). 

We summarize the existing instruction-following datasets in Table~\ref{tab:dataset_overview} of Appendix A for a better overview.


\subsubsection{Human-annotated datasets}
\label{subsubsec:human_datasets}

The conventional way to create instruction-tuning datasets is by employing human annotators, especially for those early-stage datasets, as shown in Table~\ref{tab:dataset_overview}. For example, \textsc{Public Pool of Prompts} (P3) \citep{sanh2021multitask} and \textsc{Flan} \citep{wei2021finetuned} collected multi-task instruction-tuning datasets, where they employed human expertise to design various prompts for each task. \citet{mishra2022cross} proposed \textsc{Natural Instructions}, in which they collected more than 60 NLP tasks with the corresponding human-written instructions; \citet{wang2022benchmarking} further extended this collection into a 1.6k cross-lingual tasks scale contributed by 88 NLP experts, namely \textsc{Super-Natural Instructions}.

Human-created datasets are mostly high-quality (with minimum annotation errors) but require laborious human efforts and expensive time consumption. More importantly, humans suffer from limited diversity --- it's really challenging for humans to brainstorm diverse and novel tasks; thus, the task scale of human-annotated datasets is usually limited by human annotators (e.g., expertise level and collaboration scheme of humans).

\subsubsection{LLM-synthetic datasets}
\label{subsubsec:LLM_datasets}

Since LLMs have shown their superior annotation quality on various NLP tasks~\cite{he2023annollm,pan2023gpt4reward}, tons of recent works tried to employ LLMs (e.g., ChatGPT and GPT-4) instead of humans on instruction-tuning dataset curation. For instance, \textsc{Self-Insturct}~\cite{wang2022self} and \textsc{Unnatural Instructions}~\cite{honovich2022unnatural} utilized human-annotated instructions as demonstrations to guide LLMs in devising novel tasks and increasing task diversity. \textsc{WizardLM}~\cite{xu2023wizardlm} employed an instruction evolution paradigm to increase instruction complexity. \textsc{Dynosaur}~\cite{yin2023dynosaur} repurposed existing input-output pairs in NLP datasets to stimulate new instructions and reduce annotation costs. \textsc{MUFFIN}~\cite{Lou2023MUFFIN} prompted the LLMs to gather different task instructions for the same input and obtained an impressive generalization capacity of the tuned smaller models. Besides single-turn instruction-output datasets, some works also collected multi-turn dialogue data from ShareGPT\footnote{\url{https://sharegpt.com/}}, where the instructions are created by humans (users of OpenAI API), and the responses are from LLMs~\cite{koala_blogpost_2023,vicuna2023}.

Though these LLM-synthetic datasets contained considerable noise (e.g., incoherent instructions and hallucination outputs), the diverse task distribution and model-preferred output patterns still benefit the smaller models on instruction-following, achieving comparable even better generalization performance compared with human-annotated datasets~\cite{wang2022self,wang2023far}.

In a word, the choice between human-annotated and LLM-synthetic datasets can also be regarded as a trade-off between data quality and diversity. Previous works have concluded that both factors affect the performance of the resulting models~\cite{chung2022scaling,longpre2023flan} --- mixing human and machine data can lead to better results~\cite{wang2022self,yin2023dynosaur}, while there is no concrete conclusion about which factor outweighs the other, which highly depends on the downstream tasks and application situations.

\subsection{Evaluation}

\subsubsection{Different evaluation schemes}

How to evaluate an instruction-tuned model is also a crucial topic. Most traditional NLP tasks usually have concrete criteria on the task objective, while for instruction following, the key objective is to tame the model to follow instructions --- how well the model follows instructions is highly subjective and depends on various preferences. Therefore, different works tend to utilize various evaluation strategies. In this section, we list several common evaluation settings.
 
\paragraph{Automatic metrics}

When testing the model's instruction-following performance on an evaluation dataset, if this dataset has ``ground-truth'' outputs, then a conventional criterion is to use those automatic evaluation metrics, such as \textsc{Exact-Match}~\cite{rajpurkar2016squad} and \textsc{ROUGE}~\cite{lin2004rouge}, that have been widely used for evaluating the generation models~\cite{mishra2022cross,wang2022benchmarking,wei2021finetuned,sanh2021multitask,lou2023forget,yin2022contintin}. However, this naive evaluation strategy suffers from several drawbacks: 1) it has been widely committed that the automatic generation metrics are not perfect and have significant biases (e.g., \textsc{BLUE} score has text length bias); 2) all of these metrics are used for showing how well the model's prediction aligns with pre-annotated answers, however, most real-world user tasks are highly open-ended, and there are probably no official ground-truth labels to calculate the metrics; 3) the essence of instruction following is to follow user's instructions and provide desired responses that can well address user's requirements, while automatic metrics more focus on some superficial textual patterns and lacking the reflection on how well the response satisfies the instructions.

\paragraph{Human evaluation}

A more reliable evaluation method is to employ humans to decide whether a model's response satisfies the instruction or not. For example, given a task instruction and a corresponding model output, the human evaluator should read the instruction and decide whether this model output is acceptable or not (reporting an acceptance ratio for the target model)~\cite{wang2022self,wang2023far,Lou2023MUFFIN}; or ask humans to compare two models' outputs and decide which one better satisfies the instruction (pairwise comparison between two models)~\cite{vicuna2023,koala_blogpost_2023,alpaca}. Since instructions are mostly complicated and contain a lot of explicit or implicit constraints, human evaluation is more flexible and accurate than automatic metrics in reflecting the instruction-following capacities of different models.

However, human evaluation is also much more expensive, slower than automatic evaluation, and unreproducible. Thus, most of the works only conduct human evaluation on a small subset of the whole evaluation benchmark. Meanwhile, it is mostly based on human evaluators' personal preferences and can result in high variance between different evaluators.

\paragraph{Leading LLMs as evaluators}

To address the aforementioned issues of human evaluation, recent works also tried to use LLMs (e.g., GPT-4) rather than humans to evaluate the models' instruction following capacity, such as VicunaEval~\cite{vicuna2023} and AlpacaEval~\cite{alpaca}. Nevertheless, although LLMs are cheaper and faster, they were found to have serious preference bias on some superficial textual patterns or hallucinations, e.g., GPT-4 prefers the longer texts and those responses with diverse tokens~\cite{wang2023far}. Meanwhile, only a final preference score is usually insufficient for a comprehensive evaluation.

In order to improve reliability, instead of letting LLMs simply provide a preference decision, other works tend to ask LLMs to generate comprehensive analyses besides the final decision, such as generating the error types, locations and explanations before concluding with the final scores~\cite{fernandes2023devil,xu-etal-2023-instructscore}. Some other works also predefined several explainable criterion questions for the various evaluation tasks (e.g., for an instruction ``\textit{Please generate at least 25 sentences}'', define a criterion ``\textit{is the model's generation at least 25 sentences?}''), that can be further verified by humans or LLMs easily (i.e., doing binary classification on these predefined criteria)~\cite{liu2023benchmarking,zhou2023instruction}. \citet{saha2023branch} also asked LLMs to first generate the criteria questions automatically according to the instructions and then evaluate the model's response.

\subsubsection{Two branches of evaluation}
Despite the various evaluation choices in the instruction following, they can be summarized into two branches from our view.

\paragraph{Task-centric evaluation} Most evaluation datasets in this branch are based on conventional multi-task learning, where the evaluation tasks are mostly traditional NLP tasks, such as natural language inference~\cite{wei2021finetuned,sanh2021multitask}. This branch aims to test LLMs' instruction-following and problem-solving capacity, and the main criterion here is whether the models can correctly solve the given textual task. Therefore, most of the evaluation settings in this branch adopt conventional automatic metrics to reflect the task ground-truth label alignment. Representative benchmarks are MMLU~\cite{hendrycks2020measuring}, BBH~\cite{suzgun2022challenging}, SuperNI-Test~\cite{wang2022benchmarking}, T0-Eval~\cite{sanh2021multitask}, InstructEval~\cite{chia2023instructeval}, etc.

\paragraph{Human-centric evaluation} The evaluation instructions in this setting are user-oriented or dialogue-like user queries, mainly used to test how well the models' responses align with human preference, especially for the safety and usefulness of the responses (e.g., harmlessness and honesty). Unlike the task-centric evaluation, human-centric evaluation cares less about the ground-truth labels since most user tasks are open-ended. Thus, this evaluation setting is more subjective and requires more high-level human or LLM efforts. Representative benchmarks are AlpacaFarm~\cite{dubois2023alpacafarm}, VicunaEval~\cite{vicuna2023}, HHH~\cite{bai2022constitutional}, etc.

To our knowledge, since instruction following is a pretty wide topic that can be related to various downstream tasks and real-world scenarios, there is still a lack of a comprehensive evaluation setting that can be applied to all of the target scenarios. A more practical choice is to adopt different evaluation settings according to the objectives of different works (i.e., task-centric or human-centric).

\section{Factors that Influence Instruction Following Performance}
\label{sec:analysis}

Instruction following is proven to be effective in a lot of few/zero-shot NLP tasks, but how to explain the impressive performance of instruction? And which aspects make a successful instruction following procedure? We categorize the factors affecting instruction following performance into five dimensions: \textit{model}, \textit{instruction}, \textit{demonstration}, \textit{model-instruction interaction}, and \textit{dataset}. Table~\ref{tab:takeways} displays a roadmap for this section, where we also conclude the takeaways to make it easy to refer to.

\begin{table}[t]
 \setlength{\belowcaptionskip}{-10pt}
 \setlength{\abovecaptionskip}{5pt}
    \centering
    \small
    \resizebox{0.99\linewidth}{!}{
    \begin{tblr}
    {
        width=\linewidth, 
        colspec = 
            {
            X[l]
            },
        rowspec = 
            {
            |[2pt,MyBlue]Q[b]
            |[1pt,MyBlue]
            Q[m]Q[m]
            |[0.8pt,MyBlue,dashed]
            Q[m]Q[m]
            |[0.8pt,MyBlue,dashed]
            Q[m]Q[m]
            |[0.8pt,MyBlue,dashed]
            Q[m]Q[m]
            |[0.8pt,MyBlue,dashed]
            Q[m]Q[m]
            |[2pt,MyBlue]
            },
        row{1} = {bg=MyGrey!15, font=\bfseries},,
        rowhead = 1,
        hspan = minimal,
    }
    \SetCell{c} Recipes for Instruction Following \\ 
    \colorbox{MyPink!13}{\textrm{{\em Model-related Factors}}} (\cref{subsec:modelfactors}) \\
    \begin{varwidth}[t]{\linewidth}
    \begin{itemize}[topsep=0pt,parsep=0pt]
        \item Instruction-tuned LLMs $>$ Vanilla LLMs.
        \item Instruction following tames LLMs to be more safe, robust, and user-friendly. 
        \item Larger LLMs benefit more from instruction following. \strut
    \end{itemize}
    \end{varwidth} \\
    
    \colorbox{MyPink!13}{\textrm{{\em Instruction-related Factors}}} (\cref{subsec:instruct_facor}) \\
    \begin{varwidth}[t]{\linewidth}
    \begin{itemize}[topsep=0pt,parsep=0pt]
        \item Rewriting your instruction with several epochs before it works.
        \item Keep instruction paradigm consistent during training and testing (e.g., abstractiveness).
        \item Design multiple instructions for one task in different wordings and perspectives. 
        \item Feeling exhausted about promoting diversity? Resort to the LLMs! 
        \item Few-shot demonstrations are useful in most cases. \strut
    \end{itemize}
    \end{varwidth} \\

    \colorbox{MyPink!13}{\textrm{{\em Demonstration-related Factors}}}(\cref{subsec:demon_facor}) \\
    \begin{varwidth}[t]{\linewidth}
    \begin{itemize}[topsep=0pt,parsep=0pt]
        \item The choice of your few-shot examples matters a lot!
        \item Sort your examples in a decent order.
        \item Enhance your examples with step-by-step reasoning explanation.
        \item Let your model exploit the input-output mapping from the examples. \strut
    \end{itemize}
    \end{varwidth} \\

    \colorbox{MyPink!13}{\textrm{{\em Model-Instruction Alignment}}} (\cref{subsec:model_instruct_align}) \\
    \begin{varwidth}[t]{\linewidth}
    \begin{itemize}[topsep=0pt,parsep=0pt]
        \item Better design your instructions in a model's language (e.g., conforming to the pertaining objectives). \strut
    \end{itemize}
    \end{varwidth} \\

    \colorbox{MyPink!13}{\textrm{{\em Data-wise Factors}}} (\cref{subsec:task_scale}) \\
    \begin{varwidth}[t]{\linewidth}
    \begin{itemize}[topsep=0pt,parsep=0pt]
        \item Try to tune LLMs on more diverse tasks. \strut
    \end{itemize}
    \end{varwidth} \\

    \end{tblr}
    }
    \caption{The takeaways. We summarize some high-level suggestions for successful instruction following.}
    \label{tab:takeways}
\end{table}

\subsection{Model-related Factors}
\label{subsec:modelfactors}

\subsubsection{Update model or not}

As shown in Figure~\ref{fig:two_paradigms}~(b), to drive LLMs to understand and follow task instructions more smoothly, a widely-adopted practice is fine-tuning LLMs on multi-task datasets, where each task input is equipped with a task instruction. This procedure is also well-known as ``instruction tuning''. A lot of works demonstrated that instruction-tuned LLMs could better follow the instructions of unseen tasks compared with frozen LLMs~\cite{wei2021finetuned,sanh2021multitask}.

Besides the performance gains on unseen tasks, instruction tuning has many other benefits, such as learning faster on the downstream tasks~\cite{longpre2023flan,gupta2023instruction}, being more robust to the tiny instruction perturbations (e.g., paraphrasing)~\cite{weller2020learning,sanh2021multitask,gu2022robustness}, becoming more user-friendly~\cite{chung2022scaling}, and being better at following soft instructions~\cite{wei2021finetuned}, etc. 



\begin{figure*}[t!]
\centering
\includegraphics[width=0.99\textwidth, trim=10 25 0 0]{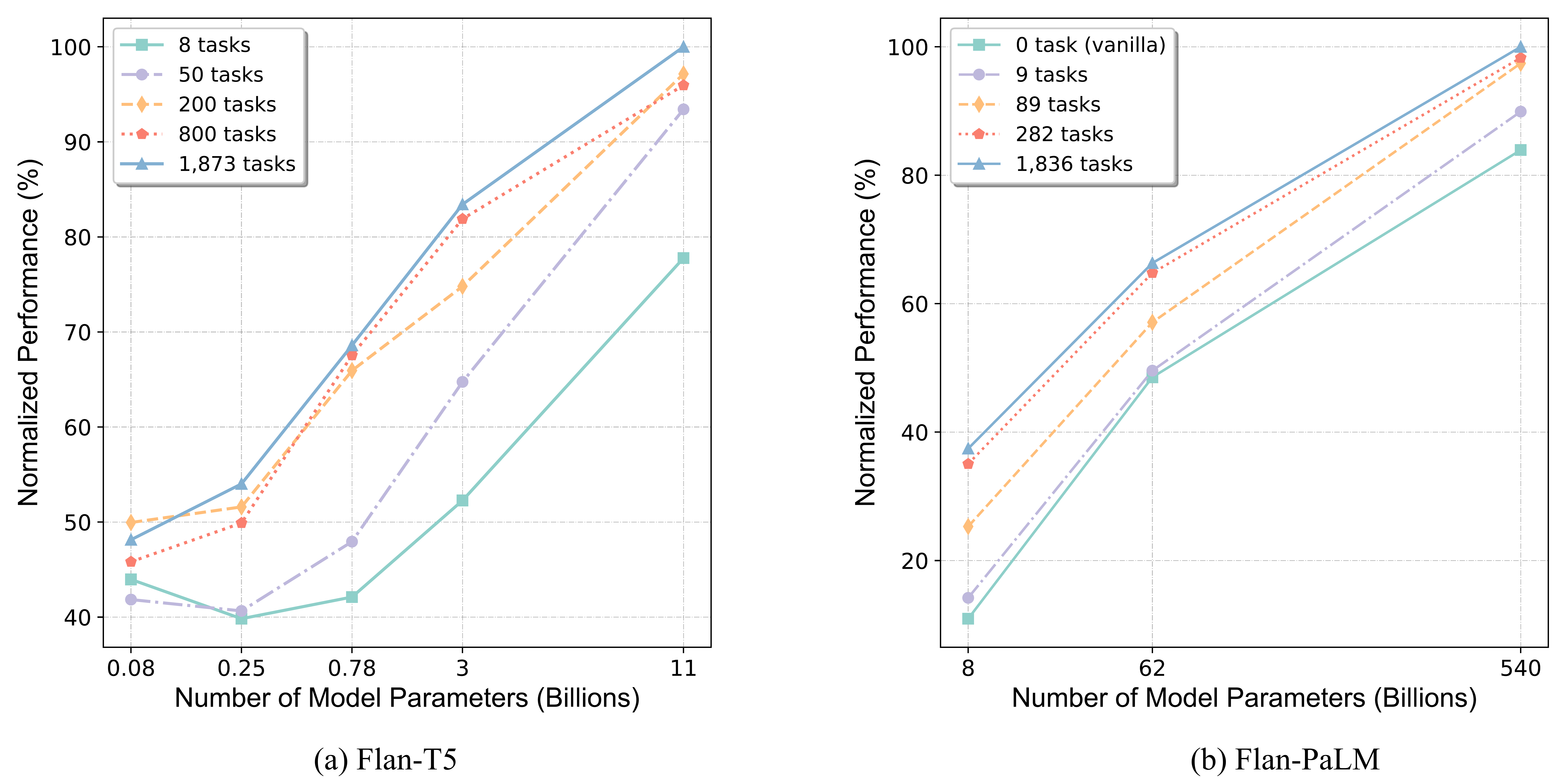}
\caption{The scaling trends of instruction following, including scaling model size and task numbers. We report the cross-task generalization performances of two widely-adopted instruction-tuned LLMs, namely Flan-T5 and Flan-PaLM~\cite{chung2022scaling}, where the source scores mainly come from~\cite{wei2021finetuned,chung2022scaling,longpre2023flan}. It is worth noting that different papers may utilize distinct evaluation benchmarks with various metrics. To clearly summarize the scaling trends, instead of simply copying the original scores, we report the \textit{normalized performances} in each figure (that's why the highest performance of each figure can reach 100\%).}
\label{fig:scaling}
\end{figure*}



\subsubsection{Model scale}

Recent works demonstrated that the model scale significantly impacts the generalization performance of instruction following~\citeiter{chung2022scaling,longpre2023flan,wang2023far}. As shown in Figure~\ref{fig:scaling}, the generalization performances of each model consistently increase when scaling up the model size. More interestingly, when the model scale is large enough, even vanilla LLMs can significantly outperform smaller LLMs tuned on extensive tasks (see Flan-PaLM; vanilla 540B $>$ 8B + 1836 tasks), which probably implies that the benefits of scaling up the model size can outweigh dataset scaling.

However, the super-large model scale is usually unaffordable for most research groups, and it also leads to huge carbon emissions, making it unrealistic in most real-world scenarios~\cite{strubell-etal-2019-energy,schick-schutze-2021-just}.
Accordingly, recent works began to investigate a more efficient way to address the model scale problem, e.g., the parameter-efficient fine-tuning~\cite{hu2021lora,liu2022few,lialin2023scaling,jang2023exploring}.
%

\subsection{Instruction-related Factors}
\label{subsec:instruct_facor}

\subsubsection{Instruction engineering}

A common problem in instruction following is that the pre-trained models are usually sensitive to some subtle modifications in the instruction~\cite{weller2020learning,efrat2020turking,bach2022promptsource,mishra-etal-2022-reframing,gu2022robustness} --- even a minor edition on instruction, such as paraphrasing or word replacement, can lead to huge performance variance. Therefore, modifying the wording of instruction before usage, namely instruction engineering, is critical for the models' performance.

One straightforward solution is to manually rewrite the instruction, i.e., human instruction engineering. When humans perform instruction engineering, the criteria of rewriting are mostly human intuition. For example, \citet{mishra-etal-2022-reframing} conducted error cases analysis on GPT's instruction-following outputs. Accordingly, they designed several empirical rules on instruction writing and ``reframed'' the instructions. All of these proposed rules are based on human intuition, e.g., itemizing instructions and task decomposition. In order to avoid the preference bias introduced by a small group of humans, \citet{bach2022promptsource} proposed community-driven instruction engineering, where they collected instructions created by various NLP experts with different writing styles, diversifying the choices of instructions. 
However, human instruction engineering is time-consuming and expensive. Moreover, the human intuition on instruction designing might be subjective and sometimes sub-optimal for the models. 

To this end, automatic instruction engineering tries to let the model figure out the better instructions automatically. \citet{prasad2022grips} proposed an edition-based method to automatically modify the instruction. For each iteration, they edited the instruction in phrase level to generate multiple candidates, and then employed the target model to predict the scores of the
different candidates by using a small labeled set (i.e., calculating the ground-truth entropy and accuracy). In doing this, \citet{prasad2022grips} achieved better performance compared with those manually-reframed instructions~\cite{mishra-etal-2022-reframing}. Besides using the ground-truth score, \citet{gonen2022demystifying} utilized the model's prediction likelihood as feedback to select instruction candidates, which even doesn't require any labeled instances. \citet{deng-etal-2022-rlprompt} further proposed a reinforcement learning framework to conduct instruction engineering. Despite the superior performance, the obvious drawback of automatic instruction engineering is the poor explainability, where the resulting instructions mostly violate human intuition (e.g., some task-irrelevant sentences)~\cite{khashabi2022prompt,prasad2022grips}, which is similar to the soft instruction.

In a word, instruction engineering is a trade-off procedure --- lower explainability is the tax of better performances. Meanwhile, instruction engineering is a highly empirical subject, and there are no gold-standard rules/methods on it --- different models and tasks might require totally different instruction designing. Hence, we highly recommend the community release the accompanying instruction manuals when releasing their instruction-tuned models, thus ensuring stable and expected model behaviours (e.g., OpenAI's cooking book\footnote{\url{https://platform.openai.com/docs/guides/prompt-engineering/six-strategies-for-getting-better-results}}).

\subsubsection{Instruction consistency}

This factor considers the \textit{instructions across the training tasks and test tasks}. Keeping the instruction paradigm (e.g., abstractiveness) consistent is crucial in instruction following. \citet{wei2021finetuned} first investigated the performance impact of changing the instruction paradigm. They found that  LLMs tuned on short instructions (i.e., task names) cannot generalize to  longer sentence-style instructions (\texttt{short $\not\Rightarrow$ long}). Similarly, \citet{gu2022robustness} observed the performance dropping when changing paragraph-style instructions to shorter sentence-style instructions at the test phase (\texttt{long $\not\Rightarrow$ short}), further indicating the importance of instruction consistency. 

Besides discrete instruction, maintaining the instruction paradigm is also critical for soft instruction, i.e., keeping the same-size prefix embedding when testing on unseen tasks~\cite{xu2022zeroprompt}. 
%
Interestingly, similar results were also found in the few-shot demonstrations (i.e., in-context learning), where the combination of input-output pairs or the number of demonstrations cannot be changed during training and evaluation~\cite{min-etal-2022-metaicl,min2022rethinking,iyer2022opt}.
These phenomena raise a concern: although instruction-tuned LLMs are robust to tiny perturbations of instructions, \textit{they are vulnerable when facing more significant alterations, which is far behind human-level generalization}.

\subsubsection{Instruction diversity}

To further improve the robustness of LLMs, especially when facing significant alterations of instruction paradigms, people try to promote instruction diversity during the \textit{training phase}---for the same training task, writing multiple instructions in different textual expressions (e.g., different wordings and lengths), then training LLMs on the mixture of diverse instructions. Notably, \citet{sanh2021multitask} showed that adopting instructions with diverse writing styles not only improved the model generalization but also compensated for the limited model scale to some extent.


Nevertheless, manually crafting instructions with diversity is expensive and usually hard to achieve due to the human annotation bias~\cite{huynh2021survey,parmar2022don}. Owing to the excellent annotation quality of LLMs~\cite{he2023annollm,pan2023gpt4reward}, a considerable number of works began to employ models to compose innovative instructions~\cite{zhang2020analogous,zhang2021learning,honovich2022instruction}. Although the model-generated instructions have been proven to contain more noise, benefiting from the diverse syntax structures~\cite{kitaev2018constituency}, these instructions could still show complementary effects with the human-written instructions~\cite{wang2022self}. More interestingly, \citet{Lou2023MUFFIN} proposed a new instruction-following dataset paradigm, where they employed LLMs to synthesize diverse task instructions for each input. Benefiting from this paradigm, the tuned LMs were forced to focus more on the instruction than the task input, achieving promising instruction-following performance.
All of these results may imply the profitability of instruction diversity, \textit{even at the expense of the correctness of instructions}. 

\subsubsection{Add demonstrations or not}

Demonstrations, i.e., a couple of input-output examples, have been shown to be critical for the expressiveness of task instructions. 
For example, existing works found that adding a few positive demonstrations in the textual instructions could result in a significant performance improvement on the unseen tasks~\cite{yin2022contintin,deb2022boosting}, especially for the tasks occupying complex output space~\cite{mishra2022cross,wang2022benchmarking}. Surprisingly, \citet{gu2022robustness} further found that 
models highly relied on few-shot demonstrations and even abandoned other useful resources (e.g., detailed task definition) when demonstrations were available.
%
This prominence is perhaps because the LLMs prefer to exploit the more superficial patterns of the demonstrations rather than the other complex textual expressions~\cite{min2022rethinking}. In other words, at present, a comprehensive framework for accurately encoding pure instructions in the absence of demonstrations or task scaling remains elusive~\cite{lou2023forget}.

\subsection{Demonstration-related Factors}
\label{subsec:demon_facor}


Since few-shot demonstrations can impact the model's instruction following performance a lot, recent studies investigated different factors in the demonstrations, which can further enhance the model's demonstration learning efficiency.

\subsubsection{The selection of demonstrations}

Given an unlabeled test instance (i.e., input-only instance waiting for the answer from the model), and a pool of labeled training instances (i.e., input-output pairs), how to select the better demonstrations from this pool for the test instance is a fundamental question for in-context learning. 

\citet{liu2022makes} proposed an unsupervised demonstration selection strategy, where they utilized \(k\)NN (\(k\) Nearest Neighbors) to retrieve the demonstrations with the closed embedding distance as the test instance. The key step in the clustering-based selection methods is the distance metrics, such as L2 distance, cosine-similarity or mutual information~\cite{sorensen2022information}. In addition to the clustering-based methods, another branch of methods used the output score of models as the selection criterion~\cite{gonen2022demystifying,wu2022self,li2023finding}. For example, \citet{nguyen2023context} tried to select a subset $A$ from the training pool as the demonstrations by measuring the model's average performance variance between $A$ and the complement set $\overline{A}$.

Beyond the above unsupervised or weak-supervised selection strategies, some other works also employed supervised methods. \citet{wang2023large} regarded the LMs as implicit topic models, where the LMs can generate meaningful concept representation based on the few-shot demonstrations. By training the topic models, they selected demonstrations that could maximize the likelihood of the given concept. Meanwhile, \citet{zhang2022active} regarded the demonstration selection as a Markov decision process~\cite{bellman1957markovian} and proposed a reinforcement learning model via Q-learning~\cite{jang2019q}.


\subsubsection{The order of demonstrations}

Even with the same set of demonstrations, the difference in the example order can also impact the model's in-context learning performance. \citet{zhao2021calibrate} emphasized that the GPT-3 is sensitive to the order of the demonstrations, and they conjectured that this sensitivity potentially comes from \textit{Recency Bias} --- the tendency to repeat answers that appear towards the end of the prompt. \citet{lu2022fantastically} further conducted comprehensive experiments and found that, along with GPT-3, various models suffer from order sensitivity. 

To this end, recent works proposed several methods to sort a ``suitable'' example order for the LMs. For example, based on the \textit{Recency Bias}, \citet{liu2022makes} calculated the embedding similarity between the demonstrations and the target input, those more similar examples were put closer (right more) to the input. \citet{lu2022fantastically} proposed several entropy-based metrics to search for the best demonstration order.

\subsubsection{Reasoning steps augmentation} 

Beyond the standard input-by-output demonstrations, augmenting in-context examples with reasoning steps is found helpful for the model's performance, especially for the super-large models.

\citet{wei2022chain} proposed chain-of-thoughts
(CoT), where they inserted some human-written intermediate reasoning
steps (i.e., rationale) between input and output of in-context demonstration. By doing so, when predicting the target output, the models can generate intermediate reasoning steps as well, thus enhancing the performance on reasoning tasks (e.g., math word problems) and the explainability of LMs. In addition to the human-written CoT, \citet{xu2023small} also found that the CoT synthesized by larger models can assist the smaller models.
Based on the promising results of adopting CoT, more advanced variations were proposed for more accurate reasoning, such as program-of-thoughts~(PoT)~\cite{chen2022program}, tree-of-thoughts~(ToT)~\cite{yao2023tree}, graph-of-thoughts~(GoT)~\cite{besta2023graph}, and CoT with self-consistency decoding augmentation~\cite{wang2022selfcons}.

However, similar to the demonstration sensitivity, different CoT writing styles can also result in performance variance. Therefore, in contrast to the human-craft CoT (i.e., few-shot CoT), \citet{zhang2022automatic} proposed Auto-CoT (i.e., zero-shot CoT), where they added a ``Let’s
think step by step'' into the prompt and let the models generate CoTs themselves. Afterwards, more and more variations of Auto-CoT were proposed to address more complicated reasoning tasks. For example, Self-Ask~\cite{press2022measuring} asked the model to first generate several questions regarding the input and then answer these questions by the model itself --- these self-generated contexts were further used as the reasoning rationales to help answer the original input. Similarly,  Least-to-Most~\cite{zhou2022least} asked the model to decompose an origin complex input into several sub-questions and answer them subsequently, which can be used as the rationales as well. 

\subsubsection{Emphasizing input-output mappings}
For in-context learning, the model usually cannot directly ``learn'' the input-output mapping from the given examples because there is no parameter update for the models. Therefore, one issue of in-context learning is that, when conducting instruction following, the demonstrations are not necessarily needed for the model to solve the task (i.e., even without the few-shot demonstrations, the model can still make predictions). \citet{min2022rethinking} also found that the model is more likely to ``copy'' the output candidate from the demonstrations, instead of truly learning the underlying mapping.

To this end, \citet{wei2023symbol}~proposed symbol tuning. Different from conventional instruction following, which tunes the models to follow input-by-output demonstrations to complete the target input, symbol tuning uses some unrelated symbols to replace the origin outputs of the demonstrations. For example, the origin output space of the demonstrations might be ``positive'' and ``negative''; symbol tuning uses ``Foo'' and ``Bar'' instead. After losing the semantics of the output spaces, there are no prior label biases~\cite{zhao2021calibrate} for the models to rely on to make the final prediction, so the models are forced to figure out the input-output mapping in the context.

\subsection{Model-Instruction Alignment}
\label{subsec:model_instruct_align}

This factor refers to making the procedure of instruction following better conform to the \textit{preference} of LLMs.
One aspect is the training objective. Since the current instruction following paradigm mainly employs the LLMs as the system backbone, one of the potential explanations for why \plminstruction~(i.e., prompt) can work is that prompt aligns well with the pretraining objective---language modeling---and activates the task-specific knowledge of the LLMs. Some existing works demonstrated the importance of conforming to the pretraining objective of LLMs when doing instruction following~\cite{schick-schutze-2021-just,tay2022unifying}, such as recalling language modeling objectives in fine-tuning phase~\cite{iyer2022opt}. 
%
Another aspect of model preference alignment is the way of designing instructions: that is, converting the instructions into model-oriented styles~\cite{deng-etal-2022-rlprompt}. For example, using soft instructions (i.e., continuous embedding) instead of human-understandable discrete instructions~\cite{lester2021power,liu2021gpt,ye2022retrieval}. It is consistent with the empirical guidelines established in the field of prompt engineering, which emphasize the significance of model-oriented prompt design.\footnote{Using prefix prompts for the auto-regressive LMs, while using cloze prompts for the masked LMs~\cite{liu2023pre}.} Despite the performance profits, it is still controversial whether it is worthwhile to convert the original human-oriented instructions into a LLM-oriented style, because it always impairs the interpretability of instructions and is highly contrary to human intuition~\cite{khashabi2022prompt,webson-pavlick-2022-prompt,prasad2022grips}.

\subsection{Data-wise Factor: Task Scale}
\label{subsec:task_scale}

The task scale often refers to the number of different training task categories in the dataset. Since ``data-wise factor'' also includes the scale of training instances, \citet{wang2022benchmarking} investigated the impact of both task and instance scales. They found that instance scale (fixed task number, increasing the number of instances per task) can only bring a limited performance boost, while task scale is the key factor for instruction following, in line with the observations of other works~\cite{wei2021finetuned,chung2022scaling}. As illustrated in Figure~\ref{fig:scaling}, the same-size model with more tuning tasks usually gains better performance. However, the performance improvement of scaling up tasks is unstable, especially when the model size is too small (e.g., 0.08B Flan-T5). This phenomenon aligns with the discussion in \cref{subsec:modelfactors}, we can draw a similar conclusion here: \textit{the profits of the task scale are highly governed by the model scale.}

\subsection{Main Takeaway: Dual-Track Scaling}

Among all the factors discussed in this section, scaling is arguably the core factor that leads to the success of instruction following. Prior to LLM-based instruction following, scaling was mainly for deep learning models: from single-layer neural nets to multi-layer perceptions, from convolutional/recurrent neural networks to deep-layer transformers~\cite{hochreiter1997long,lecun1998gradient,vaswani2017attention,devlin2018bert}. Along with the pretraining of massive raw text data, the ever-increasing models are expected to have encoded a vast amount of generic-purpose knowledge~\cite{zhou2023lima}. In the era of instruction following, where the community is more interested in cross-task generalization, merely scaling LLMs seems not enough. Thus, researchers take a parallel scaling: to collect more and more training tasks and labeled examples for each. We interpret this as a \texttt{dual-track scaling}. Overall, this dual-track scaling jointly seeks supervision to solve new tasks---the supervision either comes from LLMs' pretraining or substantial training tasks. Despite its progress, some notable challenges remain in this area, which we will discuss in the next section.

%

\section{Challenges and Future Directions}
\label{sec:challenges}
Despite all the aforementioned benefits of instruction, tons of under-explored challenges remain in this area. In this section, we list several challenges related to the instruction following, which are worthwhile for future research to investigate.

\subsection{The Tax of Instruction Alignment}

The instruction following aims at taming the models to better assist humans in real-world tasks; therefore, in addition to pursuing ultimate performance, inference-time safety is also a crucial aspect for the instruction-tuned models (i.e., instruction alignment). \citet{ouyang2022training} defined ``alignment'' with three criteria --- \textit{Helpful}, \textit{Honest}, and \textit{Harmless} (HHH), which has been widely considered by the previous instruction tuning models and datasets~\cite{bai2022constitutional,yin2023dynosaur,wang2022self,Lou2023MUFFIN}. However, alignment can also bring ``tax'' to the instruction-tuned models. For example, \citet{bekbayev2023poison} found that the well-aligned answers provided in the instruction following datasets can drop the model's performance a lot on various task benchmarks. This implies a trade-off between performance and safety for instruction following, which requires careful consideration.

\subsection{Learning Negated Information}

Negation is the common linguistic property and has been found to be crucial for various NLP tasks, e.g.,  NLI \cite{naik2018stress,kassner2020negated}. Specific to instruction following, negation denotes any \textit{things-to-avoid} information of in-context instructions, including negated requirements (e.g., ``\texttt{avoid using stop words}'') and negative demonstrations (i.e., some wrong examples). Although humans can learn a lot from the negation~\cite{dudschig2018does}, existing works found LLMs often fail to follow the negated instructions; some negations can even drop models' performance~\cite{li2022maqa,jang2022can,mishra-etal-2022-reframing}.   

Since negation has increasingly become a challenge in instruction following, we provide several hints to inspire future work. One potential solution is unlikelihood training~\cite{hosseini2021understanding,ye2022guess}, which trains the LLMs to minimize the ground truth probability when negated instructions are conditioned. Besides, \citet{yin2022contintin} proposed pretraining the LMs on the negative demonstrations with maximizing likelihood objective to exploit the useful information in the negation. Some other methods, such as contrast-consistent projection~\cite{burns2022discovering} and n-gram representations~\cite{sun-lu-2022-implicit}, also provided insights into tackling this problem.

\subsection{Adversarial Instruction Attacks}

Though most of the instruction-tuned LMs can align well with human preferences and provide harmless responses, recent works found that they could easily be attacked --- the model's response can be manipulated by using simple prompting strategies. \citet{kang2023exploiting} designed several prompts to trigger the LLMs to generate malicious content. For example, instead of directly providing malicious instruction with obviously harmful intentions, they split the instruction into several pieces (each piece itself doesn't trigger the LLMs' defence mechanism). In doing this, those powerful preference-aligned LLMs, such as ChatGPT and InstructGPT, were successfully fooled and generated harmful content. \citet{li2023you} also found that the retrieval-augmented generation models can be easily attacked by injecting adversarial questions into the retrieved context. Besides attacking the instruction-tuned LLMs, \citet{wan2023poisoning} concluded that LLMs can also be attacked during instruction following. Based on the clean instances, they automatically created a few poisoned examples to train the LLMs and found that the resulting LLMs could be manipulated by using some trigger words.

Since instruction-tuned LLMs have been applied to various real-world scenarios, such as web agents and search engines~\cite{deng2023mind2web,xie2023adaptive}, the safety of LLMs' generations is becoming more urgent. Simply conducting preference alignment or content filtering seems to be insufficient, especially for those super-strong LLMs. Thus, developing efficient defence methods is necessary for the current instruction-tuned models. Meanwhile, further deep analyses of LLMs' vulnerability are also critical, potentially providing more insights into the defence.

\subsection{Explainability of Instruction Following}

As we have mentioned in~\cref{sec:analysis}, to achieve a promising cross-task performance, one of the critical factors is to convert the \humaninstruction~into \plminstruction, i.e., making the instructions conform to the model's preference.
Numerous previous works have verified the effectiveness of catering to the model's preference in designing instructions, e.g., using the model's perplexity in choosing appropriate instructions~\cite{gonen2022demystifying}.
Despite the performance gains, the resulting instructions consistently violate human intuitions and show worrying reliability, such as some semantically incoherent, task-irrelevant, or even misleading instructions~\cite{khashabi2022prompt,prasad2022grips}.
\textit{These results prove the conflict between performance profits and the human interpretability of instructions, which is tricky to trade-off.}

Although \citet{mishra-etal-2022-reframing} demonstrated that it is possible to maintain both the faithfulness and effectiveness of instructions, manual rewriting requires laborious human efforts. Therefore, one of the future trends is to investigate how to automatically rephrase the instructions, in a way that matches both human and model preferences.


\subsection{Learning to Follow Instruction rather than Merely Generating $\textsc{Y}$}

Multi-task instruction following is becoming a fundamental practice in the current instruction following paradigm. However, there are two issues in such a learning paradigm: (i) it relies on training on massive labeled examples to learn the instructions, which is still expensive and unrealistic for using large-scale LLMs; (ii) although the ultimate goal of instruction following is learning to follow instructions by observing various training tasks, the current training objective is still the conventional maximum likelihood of reference outputs. This implicit instruction following objective can lead to sub-optimal optimization (i.e., LLMs can  learn to generate $\textsc{Y}$ for $\textsc{X}$ without really understanding the meaning of instructions $\textsc{I}$).

To this end, one desired future direction is to evolve a new learning objective to help LLMs explicitly learn to follow instructions, which might alleviate the reliance on large-scale labeled instances. 
Moreover, a more ambitious and challenging idea is to drive the system to follow instructions without additional tuning on the labeled examples of any specific tasks~\cite{ye2023context}, which is somehow similar to semantic parser-based paradigm (\cref{sec:modeling}).

\subsection{Multi-Lingual Instruction Following}

Intuitively, the instruction following is language-agnostic capacity for the language models, which means that it is also possible for the multi-lingual language models to follow the same semantic instructions with different languages. For example, \citet{kew2023turning} found LLMs tuned with more than three languages exhibit stronger instruction following capacity, implying the benefits of multi-lingual instruction tuning. Unfortunately, most of the current open-sourced instruction following datasets and foundation models are English-centric (as shown in Table~\ref{tab:dataset_overview}). Therefore, the release of high-quality multi-lingual instruction tuning datasets (with pair translation) should be valuable for future research, as also mentioned by \citet{peng2023gpt4llm}.

\section{Instruction-related Applications}
\label{sec:app}

In addition to the main body of our paper, we also survey some popular instruction-related application directions to inspire future board-wide utilization for instruction following.

\subsection{Human-Computer Interaction}
\label{subsec:HCI}

Textual instructions can be naturally regarded as a human-computer interaction method. Numerous previous works employed natural language instructions to guide the computer to perform various real-world tasks. 

For the non-NLP (multi-modal) tasks, most focused on environment-grounded language learning, i.e., driving the agent to associate natural language instructions with the environments and make corresponding reactions, such as selecting mentioned objects from an image/video~\cite{matuszek2012joint,krishnamurthy2013jointly,puig2018virtualhome}, following navigational instructions to move the agent~\cite{tellex2011approaching,kim2012unsupervised,chen2012fast,artzi2013weakly,bisk2016natural}, plotting corresponding traces on a map~\cite{vogel2010learning,chen2011learning}, playing soccer/card games based on given rules~\cite{kuhlmann2004guiding,eisenstein2009reading,branavan2011learning,babecs2012learning,goldwasser2014learning}, generating real-time sports broadcast~\cite{chen2008learning,liang2009learning}, controlling software~\cite{branavan2010reading}, and querying external databases~\cite{clarke2010driving}, etc.
Meanwhile, instructions are also widely adapted to help communicate with the system in solving NLP tasks, e.g., following instructions to manipulate strings~\cite{gaddy2019pre}, classifying emails based on the given explanations~\cite{srivastava2017joint,srivastava2018zero}, and text-to-code generation~\cite{acquaviva2021communicating}.

Recently, a growing body of research tended to design the human-computer communication procedure in an \textit{iterative} and \textit{modular} manner~\cite{dwivedi2022editeval,chakrabarty2022help}. For example, \citet{li2020interactive} built a system to help the users tackle daily missions (e.g., ordering coffee or requesting Uber). Benefiting from a user-friendly graphical interface, the system can iteratively ask questions about the tasks, and users can continually refine their instructions to avoid unclear descriptions or vague concepts. 
%
As it is usually hard for non-expert users to write sufficient instructions in one shot, adapting an iterative and modular paradigm in designing instruction-based AI systems can help guide the users to enrich the task instruction step by step. Thus, this paradigm efficiently relieves the thinking demands of users and leads to a more user-oriented system~\cite{mishra2022help}. Due to its practical values, we emphasize the importance of this branch of work in this paper.

\subsection{Data and Feature Augmentation}

Task instructions are regarded as indirect supervision resources where sometimes superficial and assertive rules are embedded. These rules are also known as \textit{labeling functions} that can be directly applied for annotations.\footnote{For example, if ``a very fair price'' is sentiment-positive, every sentence with a similar adj-noun collocation as ``fair price'' will be positive as well.}
Therefore, some existing works also employed the instruction as a distant supervision to perform data or feature augmentation~\cite{srivastava2018zero,hancock2018training,ye2020teaching}.
For instance, \citet{srivastava2017joint} used a semantic parser to convert natural language explanations into logical forms, and applied them on all instances in the dataset to generate additional binary features. \citet{wang2020learning} utilized the label explanations to annotate the raw corpus automatically and trained the classifier on the resulting noisy data. 

Besides the straightforward augmentation, \citet{su2022one} further used the task instruction to enrich the model representation and achieved strong cross-task generalization. Specifically, they trained an embedding model (a single encoder) on the diverse instruction datasets with contrastive learning, and then used this model to produce task-specific representations based on the instruction for the downstream unseen tasks.


\subsection{Generalist Language Models}

According to the definition of Artificial General Intelligence (AGI), the ``generalist model'' is usually a system that can be competent for different tasks and scalable in changeable contexts, which shall go far beyond the initial anticipations of its creators~\cite{wang2007introduction,goertzel2014artificial}.
While specific to the NLP domain, a generalist language model is supposed to be an excellent multi-task assistant, that is skilled in handling a variety of real-world NLP tasks and different languages, in a completely zero/few-shot manner~\cite{arivazhagan2019massively,pratap2020massively,wei2021finetuned}.
As numerous existing works demonstrated the incredible power of using instructions in cross-task generalization~\citeiter{wei2021finetuned,sanh2021multitask,mishra2022cross,wang2022benchmarking,chung2022scaling}, the instruction is likely to become a breakthrough in achieving this ultimate goal. 

Notably, the recent remarkable applications of instructions, namely InstructGPT, ChatGPT, and GPT-4, also indicated a big step towards building generalist language models. For example, during the pretraining of LLama-2, \citet{touvron2023llama} utilized the idea of context distilling to inculcate
instructions within LLMs, thus addressing the inconsistency issue of instruction following in the multi-turn dialogue situation. OpenAI GPT-series adopt RLHF to align the model's preference with human instructions, where feedback supervision plays a big role.
Although the answer to ``\textit{Is instruction or human feedback, that contributes more to the performance of ChatGPT?}'' remains ambiguous and needs further investigation, we introduce some recent works highlighting the critical role of instruction following.
For example, \citet{chung2022scaling} conducted extensive experiments to evaluate the human-preference alignments of PaLM~\cite{chowdhery2022palm}. They found that, even without any human feedback, the instruction following significantly reduced the toxicity in the open-ended generations of PaLM, such as gender and occupation bias. In addition, some other works also solely employed creative instructions instead of human feedback and achieved notable cross-task results~\cite{bai2022constitutional,honovich2022unnatural,wang2022self}. Furthermore, as the knowledge conflict problem of LLMs has a significant impact on the applications of instruction-tuned models~\cite{xie2023adaptive}, in order to make the LLMs more generalist and useful in the real world, recent works also utilized the idea of the instruction following to enhance the retrieval-augmented language models, and vice versa, improve the instructions by adopting retrieved knowledge~\cite{lin2023ra}.

\section{Conclusion}


This survey summarizes the existing literature on instruction following, providing a comprehensive overview of the field, including instruction taxonomies, modeling strategies, and key aspects of instruction utilization. It also addresses unique challenges and offers hints for future research. Unlike previous works, we go beyond the limited scope of modern instruction following---we trace the studies of instruction following back to the early stage of machine learning, and explore textual instruction as an indirect supervision for LLMs. To our knowledge, this is the first extensive survey on instruction following. Overall, we aim to offer valuable insights and inspire further in-depth research in this area.


\bibliography{anthology}
\bibliographystyle{acl_natbib}

\appendix
\section*{Appendix A. Instruction-following Datasets}

\begin{table*}[ht]
 \setlength{\belowcaptionskip}{-10pt}
 \setlength{\abovecaptionskip}{5pt}
    \centering
    \large
    \renewcommand{\arraystretch}{1.56}
    \resizebox{0.99\linewidth}{!}{
\begin{tabular}{l|c|rr|c|c}
\toprule
\multirow{2}{*}{\textbf{Datasets}} & \multirow{2}{*}{\textbf{Release Time}} & \multicolumn{2}{c|}{\textbf{Scale}}                                                             & \multirow{2}{*}{\textbf{Language}} & \multirow{2}{*}{\textbf{Annotator}}                                               \\ \cmidrule{3-4}
                                   &                                        & \multicolumn{1}{c|}{\textbf{\# of Tasks}} & \multicolumn{1}{c|}{\textbf{\#   of Instances (k)}} &                                    &                                                                                   \\ \midrule
\textbf{UnifiedQA}~\cite{khashabi-etal-2020-unifiedqa}                          & 05/2020                                 & \multicolumn{1}{r|}{46}                   & 750                                                 & \encircle[fill=myOrangev2, text=white]{monolingual}                        & \includegraphics[width=0.19in]{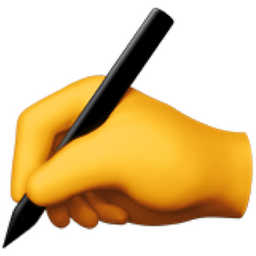} Human                                                                           \\ \hline
\textbf{CrossFit}~\cite{ye-etal-2021-crossfit}                           & 	04/2021                                       & \multicolumn{1}{r|}{159}                  & 71,000                                              & \encircle[fill=myOrangev2, text=white]{monolingual}                        & \includegraphics[width=0.19in]{picture/writing-hand.png} Human                                                                           \\ \hline
\textbf{Natural Instructions}~\cite{mishra2022cross}               &    04/2021                                    & \multicolumn{1}{r|}{61}                   & 620                                                 & \encircle[fill=myOrangev2, text=white]{monolingual}                        & \includegraphics[width=0.19in]{picture/writing-hand.png} Human                                                                           \\ \hline
\textbf{Flan 2021}~\cite{wei2021finetuned}                          &      09/2021                                  & \multicolumn{1}{r|}{62}                   & 4,400                                               & \encircle[fill=myOrangev2, text=white]{monolingual}                        & \includegraphics[width=0.19in]{picture/writing-hand.png} Human                                                                           \\ \hline
\textbf{P3}~\cite{sanh2021multitask}                                 &       10/2021                                & \multicolumn{1}{r|}{62}                   & 12,000                                              & \encircle[fill=myOrangev2, text=white]{monolingual}                        & \includegraphics[width=0.19in]{picture/writing-hand.png} Human                                                                           \\ \hline
\textbf{MetaICL}~\cite{min-etal-2022-metaicl}                            &       10/2021                                 & \multicolumn{1}{r|}{142}                  & 3,500                                               & \encircle[fill=myOrangev2, text=white]{monolingual}                        & \includegraphics[width=0.19in]{picture/writing-hand.png} Human                                                                           \\ \hline
\textbf{ExMix}~\cite{aribandiext5}                              &                   11/2021                     & \multicolumn{1}{r|}{107}                  & 500                                                 & \encircle[fill=myOrangev2, text=white]{monolingual}                        & \includegraphics[width=0.19in]{picture/writing-hand.png} Human                                                                           \\ \hline
\textbf{Super-Natural Instructions}~\cite{wang2022benchmarking}         &           04/2022                             & \multicolumn{1}{r|}{1,613}                & 5,000                                               & \encircle[fill=MyRedv2, text=white]{multilingual}                       & \includegraphics[width=0.19in]{picture/writing-hand.png} Human                                                                           \\ \hline
\textbf{GLM}~\cite{zeng2022glm}                                &               10/2022                         & \multicolumn{1}{r|}{77}                   & 12,000                                              & \encircle[fill=MyGreenv2, text=white]{bilingual}                         & \includegraphics[width=0.19in]{picture/writing-hand.png} Human                                                                           \\ \hline
\textbf{Flan 2022}~\cite{longpre2023flan}                          &           10/2022                             & \multicolumn{1}{r|}{1,836}                & 15,000                                              & \encircle[fill=MyRedv2, text=white]{multilingual}                       & \includegraphics[width=0.19in]{picture/writing-hand.png} Human                                                                           \\ \hline
\textbf{xP3}~\cite{muennighoff2022crosslingual}                                &  11/2022                                      & \multicolumn{1}{r|}{71}                   & 81,000                                              & \encircle[fill=MyRedv2, text=white]{multilingual}                       & \includegraphics[width=0.19in]{picture/writing-hand.png} Human                                                                           \\ \hline
\textbf{Unnatural Instructions}~\cite{honovich2022unnatural}             &           12/2022                             & \multicolumn{1}{r|}{117}                  & 64                                                  & \encircle[fill=myOrangev2, text=white]{monolingual}                        & \includegraphics[width=0.19in]{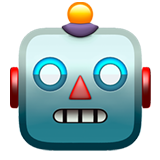} InstructGPT                                                                  \\ \hline
\textbf{Self-Instruct}~\cite{wang2022self}                      &                 12/2022                       & \multicolumn{1}{r|}{/}                    & 82                                                  & \encircle[fill=myOrangev2, text=white]{monolingual}                        & \includegraphics[width=0.19in]{picture/1f916.png} GPT-3                                                                        \\ \hline
\textbf{OPT-IML}~\cite{iyer2022opt}                           &                 12/2022                       & \multicolumn{1}{r|}{2,207}                & 18,000                                              & \encircle[fill=MyRedv2, text=white]{multilingual}                       & \includegraphics[width=0.19in]{picture/writing-hand.png} Human                                                                           \\ \hline
\textbf{Alpaca}~\cite{alpaca}                             &                  03/2023                      & \multicolumn{1}{r|}{/}                    & 52                                                  & \encircle[fill=myOrangev2, text=white]{monolingual}                        & \includegraphics[width=0.19in]{picture/1f916.png} InstructGPT                                                                  \\ \hline
\textbf{Baize}~\cite{xu2023baize}                              &              04/2023                          & \multicolumn{1}{r|}{/}                    & 100                                                 & \encircle[fill=myOrangev2, text=white]{monolingual}                        & \includegraphics[width=0.19in]{picture/1f916.png} ChatGPT                                                                      \\ \hline
\textbf{Koala}~\cite{koala_blogpost_2023}                              &        04/2023                                & \multicolumn{1}{r|}{/}                    & /                                                   & \encircle[fill=myOrangev2, text=white]{monolingual}                        & \begin{tabular}[c]{@{}c@{}}\includegraphics[width=0.19in]{picture/writing-hand.png} Human\\     \includegraphics[width=0.19in]{picture/1f916.png} ChatGPT\end{tabular}                 \\ \hline
\textbf{GPT4All}~\cite{anand2023gpt4all}                            &             04/2023                           & \multicolumn{1}{r|}{/}                    & 808                                                 & \encircle[fill=myOrangev2, text=white]{monolingual}                        & \begin{tabular}[c]{@{}c@{}}\includegraphics[width=0.19in]{picture/writing-hand.png} Human\\     \includegraphics[width=0.19in]{picture/1f916.png} ChatGPT\end{tabular}                 \\ \hline
\textbf{Alpaca-gpt4}~\cite{peng2023gpt4llm}                        &                04/2023                        & \multicolumn{1}{r|}{/}                    & 113                                                 & \encircle[fill=MyGreenv2, text=white]{bilingual}                          & \includegraphics[width=0.19in]{picture/1f916.png} GPT-4                                                                        \\ \hline
\textbf{Vicuna}~\cite{vicuna2023}                             &                  04/2023                      & \multicolumn{1}{r|}{/}                    & 76                                                  & \encircle[fill=myOrangev2, text=white]{monolingual}                        & \begin{tabular}[c]{@{}c@{}}\includegraphics[width=0.19in]{picture/writing-hand.png} Human\\      \includegraphics[width=0.19in]{picture/1f916.png} ChatGPT\end{tabular}                 \\ \hline
\textbf{Dolly}~\cite{dolly2023}                              &                 04/2023                       & \multicolumn{1}{r|}{/}                    & 15                                                  & \encircle[fill=myOrangev2, text=white]{monolingual}                        & \includegraphics[width=0.19in]{picture/writing-hand.png} Human                                                                           \\ \hline
\textbf{Oasst}~\cite{kopf2023openassistant}                              &        04/2023                                & \multicolumn{1}{r|}{/}                    & 84                                                  & \encircle[fill=MyRedv2, text=white]{multilingual}                       & \includegraphics[width=0.19in]{picture/writing-hand.png} Human                                                                           \\ \hline
\textbf{LongForm}~\cite{koksal2023longform}                           &             04/2023                           & \multicolumn{1}{r|}{/}                    & 27                                                  & \encircle[fill=myOrangev2, text=white]{monolingual}                        & \begin{tabular}[c]{@{}c@{}}\includegraphics[width=0.19in]{picture/writing-hand.png} Human\\     \includegraphics[width=0.19in]{picture/1f916.png} InstructGPT\end{tabular}             \\ \hline
\textbf{Symbolic-Instruct}~\cite{liu2023zero}                  &                  04/2023                      & \multicolumn{1}{r|}{/}                    & 796                                                 & \encircle[fill=myOrangev2, text=white]{monolingual}                        & \includegraphics[width=0.19in]{picture/writing-hand.png} Human                                                                           \\ \hline
\textbf{LaMini}~\cite{wu2023lamini}                             &                 04/2023                       & \multicolumn{1}{r|}{/}                    & 2,580                                               & \encircle[fill=myOrangev2, text=white]{monolingual}                        & \includegraphics[width=0.19in]{picture/1f916.png} ChatGPT                                                                      \\ \hline
\textbf{WizardLM}~\cite{xu2023wizardlm}                           &              04/2023                          & \multicolumn{1}{r|}{/}                    & 196                                                 & \encircle[fill=myOrangev2, text=white]{monolingual}                        & \includegraphics[width=0.19in]{picture/1f916.png} ChatGPT                                                                      \\ \hline
\textbf{COEDIT}~\cite{raheja2023coedit}                             &             05/2023                           & \multicolumn{1}{r|}{/}                    & 82                                                  & \encircle[fill=myOrangev2, text=white]{monolingual}                        & \includegraphics[width=0.19in]{picture/writing-hand.png} Human                                                                           \\ \hline
\textbf{UltraChat}~\cite{ding2023enhancing}                          &             05/2023                           & \multicolumn{1}{r|}{/}                    & 1,500                                               & \encircle[fill=myOrangev2, text=white]{monolingual}                        & \includegraphics[width=0.19in]{picture/1f916.png} ChatGPT                                                                      \\ \hline
\textbf{CoT Collection}~\cite{kim2023cot}                     &                      05/2023                  & \multicolumn{1}{r|}{1,060}                & 1,880                                               & \encircle[fill=myOrangev2, text=white]{monolingual}                        & \includegraphics[width=0.19in]{picture/1f916.png} Codex                                                                        \\ \hline
\textbf{Dynosaur}~\cite{yin2023dynosaur}                           &                05/2023                        & \multicolumn{1}{r|}{5,740}                & 801                                                 & \encircle[fill=myOrangev2, text=white]{monolingual}                        & \includegraphics[width=0.19in]{picture/1f916.png} ChatGPT                                                                      \\ \hline
\textbf{MUFFIN}~\cite{Lou2023MUFFIN}                             &                  10/2023                      & \multicolumn{1}{r|}{/}                    & 68                                                  & \encircle[fill=myOrangev2, text=white]{monolingual}                        & \begin{tabular}[c]{@{}c@{}}\includegraphics[width=0.19in]{picture/writing-hand.png} Human\\      \includegraphics[width=0.19in]{picture/1f916.png} ChatGPT\\     \includegraphics[width=0.19in]{picture/1f916.png} GPT-4\end{tabular} \\ \hline
\textbf{Dynamics-of-Instruction}~\cite{song2023dynamics}                             &                  10/2023                      & \multicolumn{1}{r|}{/}                    & 40                                                  & \encircle[fill=myOrangev2, text=white]{monolingual}                        & \includegraphics[width=0.19in]{picture/writing-hand.png} Human \\ \hline
\textbf{CoachLM}~\cite{liu2023automatic}                             &                  11/2023                      & \multicolumn{1}{r|}{/}                    & 2                                                  & \encircle[fill=myOrangev2, text=white]{monolingual}                        & \includegraphics[width=0.19in]{picture/writing-hand.png} Human \\ \hline
\textbf{DEITA}~\cite{liu2023makes}                             &                  12/2023                      & \multicolumn{1}{r|}{/}                    & 10                                                  & \encircle[fill=myOrangev2, text=white]{monolingual}                        & \includegraphics[width=0.19in]{picture/1f916.png} ChatGPT \\ \hline
\textbf{WaveCoder}~\cite{yu2023wavecoder}                             &                  12/2023                      & \multicolumn{1}{r|}{4}                    & 20                                                  & \encircle[fill=myOrangev2, text=white]{monolingual}                        & \begin{tabular}[c]{@{}c@{}}\includegraphics[width=0.19in]{picture/1f916.png} ChatGPT\\      \includegraphics[width=0.19in]{picture/1f916.png} GPT-4 \end{tabular}                 \\ 
\bottomrule
\end{tabular}
}
\caption{Instruction-tuning datasets summarization. Due to the diverse user tasks, some datasets didn't report the task scale.} 
    \label{tab:dataset_overview}
\end{table*}

\end{document}